\documentclass[conference]{IEEEtran}
\IEEEoverridecommandlockouts
\usepackage{amsmath,amssymb,amsfonts}
\usepackage{algorithmic}
\usepackage{graphicx}
\usepackage{textcomp}
\usepackage{xcolor}
\usepackage{float}
\usepackage{times}
\usepackage{listings}
\usepackage{mdframed}
\usepackage{soul}
\usepackage{inconsolata}
\usepackage{tcolorbox}
\usepackage{fancyvrb}
\usepackage{caption}
\usepackage{makecell}

\usepackage{multicol}
\usepackage{booktabs}
\usepackage{multirow}
\usepackage{color}
\usepackage{ulem}
\usepackage{bbding}
\usepackage{colortbl}
\usepackage{tikz}
\usepackage{chapterbib}    
\usepackage{kantlipsum}
\usepackage{cuted}
\let\labelindent\relax
\usepackage{enumitem}
\usepackage[figuresright]{rotating}
\usepackage{subcaption}

\usepackage{caption}
\usepackage{longtable}
\usepackage{pifont}
\newcommand{\cmark}{\ding{51}}%
\newcommand{\xmark}{\ding{55}}%

\usepackage[noadjust]{cite}

\lstdefinestyle{customstyle}{
    language=TeX, 
    basicstyle=\small\ttfamily,
    keywordstyle=\color{blue},
    commentstyle=\color{green!50!black},
    numbers=left,
    numberstyle=\tiny,
    numbersep=5pt,
    breaklines=true,
    showstringspaces=false,
    frame=none,
}

\definecolor{myhighlight}{RGB}{255, 120, 45}  
\sethlcolor{myhighlight}

\definecolor{itwasallyellow}{HTML}{F9D098}
\definecolor{darkyellow}{HTML}{915708} 

\usepackage{hyperref}
\usepackage{bbm}
\hypersetup{
    colorlinks=true,
    linkcolor=orange,
    filecolor=orange,      
    urlcolor=orange,
    citecolor=orange,
}
\usepackage[capitalise, nameinlink]{cleveref}

\usepackage[numbers, sort&compress]{natbib}
\usepackage{multicol} 
\usepackage{graphicx}

\def\BibTeX{{\rm B\kern-.05em{\sc i\kern-.025em b}\kern-.08em
    T\kern-.1667em\lower.7ex\hbox{E}\kern-.125emX}}
\begin{document}

\title{\huge{\textbf{FEAST}: A \underline{\textbf{F}}lexible M\underline{\textbf{e}}altime-\underline{\textbf{A}}ssistance  \underline{\textbf{S}}ystem \\ \underline{\textbf{T}}owards In-the-Wild Personalization}
}


\author{
\authorblockN{
Rajat Kumar Jenamani$^1$, 
Tom Silver$^1$, 
Ben Dodson$^1$, 
Shiqin Tong$^1$, 
Anthony Song$^1$,}
\authorblockN{
Yuting Yang$^2$, 
Ziang Liu$^1$, 
Benjamin Howe$^3$, 
Aimee Whitneck$^3$, 
Tapomayukh Bhattacharjee$^1$
}\vspace{-0.4cm}\\
\authorblockA{$^1$Cornell University, $^2$University of Michigan, $^3$Independent researcher}}

\twocolumn[{%
\renewcommand\twocolumn[1][]{#1}%
\maketitle


\vspace{-0.3cm}
\begin{center}
    \captionsetup{type=figure}
    \vspace{-0.4cm}
    \includegraphics[width=\textwidth]{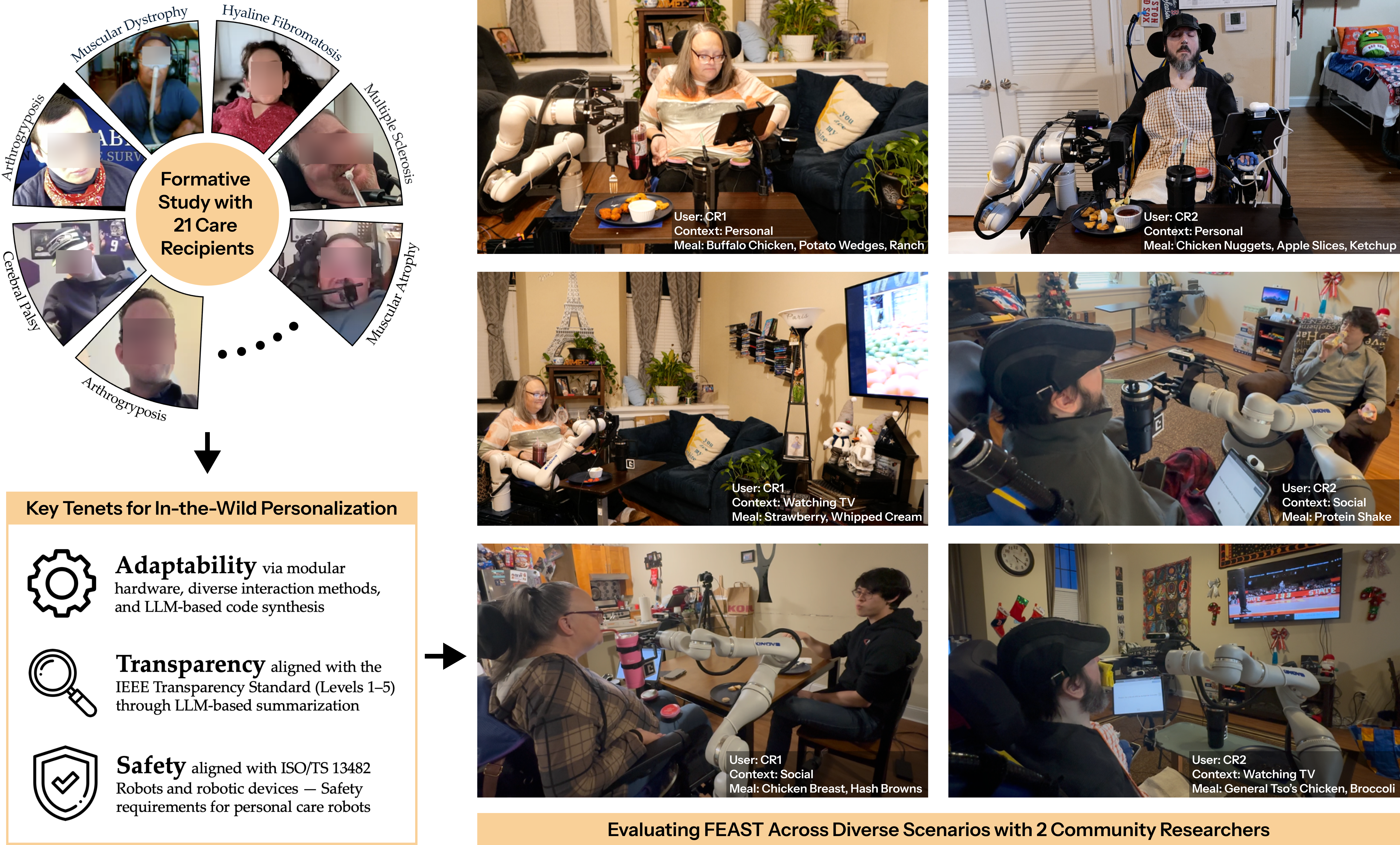}
    \captionof{figure}{Informed by a formative study with 19 care recipients and 2 community researchers, we propose FEAST: a flexible mealtime-assistance system grounded in adaptability, transparency, and safety. Through in-home evaluations, we demonstrate that FEAST can personalize to the needs of two care recipients, assisting each with three meals in diverse in-the-wild contexts.}
    \label{fig:teaser}
\end{center}
\vspace{-0.2cm}

}]

\begin{abstract}
Physical caregiving robots hold promise for improving the quality of life of millions worldwide who require assistance with feeding. However, in-home meal assistance remains challenging due to the diversity of activities (e.g., eating, drinking, mouth wiping), contexts (e.g., socializing, watching TV), food items, and user preferences that arise during deployment. In this work, we propose FEAST, a flexible mealtime-assistance system that can be personalized in-the-wild to meet the unique needs of individual care recipients. Developed in collaboration with two community researchers and informed by a formative study with a diverse group of care recipients, our system is guided by three key tenets for in-the-wild personalization: adaptability, transparency, and safety. FEAST embodies these principles through: (i) modular hardware that enables switching between assisted feeding, drinking, and mouth-wiping, (ii) diverse interaction methods, including a web interface, head gestures, and physical buttons, to accommodate diverse functional abilities and preferences, and (iii) parameterized behavior trees that can be safely and transparently adapted using a large language model. We evaluate our system based on the personalization requirements identified in our formative study, demonstrating that FEAST offers a wide range of transparent and safe adaptations and outperforms a state-of-the-art baseline limited to fixed customizations. To demonstrate real-world applicability, we conduct an in-home user study with two care recipients (who are community researchers), feeding them three meals each across three diverse scenarios. We further assess FEAST's ecological validity by evaluating with an Occupational Therapist previously unfamiliar with the system. In all cases, users successfully personalize FEAST to meet their individual needs and preferences. Supplementary materials and videos can be found at: \href{http://emprise.cs.cornell.edu/feast}{emprise.cs.cornell.edu/feast}.

\end{abstract}


\vspace{-0.3cm}
\section{Introduction}

Eating is a fundamental part of human life, deeply intertwined with identity and social interaction~\cite{nanavati2023design}. The inability to self-feed has been associated with profound emotional impacts, including feelings of shame, diminished self-esteem, and heightened anxiety or fear~\cite{mandy2018manual, jacobsson2000eatingprocess, shune2020experience}. Unfortunately, millions worldwide require assistance with feeding due to spinal cord injuries, strokes, cerebral palsy, old age, and other health conditions~\cite{taylor2018americans}. For caregivers, feeding is one of the most time-consuming Activities of Daily Living (ADLs) ~\cite{chio2006caregiver}, contributing significantly to their already substantial workload~\cite{fleming2003caregiver, zenker2017exploring}.

Robot mealtime-assistance systems have the potential to assist care recipients and improve their quality of life~\cite{brose2010role} while decreasing the physical workload on caregivers~\cite{bhattacharjee2020moreautonomy, jenamani2024bitetransfer}. 
Recent advancements have significantly improved various aspects of mealtime assistance, including food manipulation~\cite{feng2019robot, gordon2020adaptive, gordon2021leveraging, gordon2023towards, grannen2022learning, tai2023scone, liu2024imrl, sundaresan2023learning}, skill sequencing~\cite{jenamani2024flair, ha2024repeat, bhaskar2024lava}, and bite transfer~\cite{gallenberger2019transfer, belkhale2022balancing, park2020evaluation, shaikewitz2022mouth, jenamani2024bitetransfer}.
Although further progress is needed before these systems can operate long-term without expert supervision, these recent works significantly enhance their robustness and autonomy.

Building upon these efforts, this work considers the need for \emph{personalization} in mealtime assistance.
A one-size-fits-all mealtime assistance system is not enough to address the wide range of preferences, functional abilities, behaviors, and environmental contexts that vary between care recipients and between meals~\cite{madan2022sparcs}.
For example, consider the two community researchers (CRs) shown in Figure~\ref{fig:teaser}, who are both co-authors on this paper.
CR1, who has Multiple Sclerosis, prefers to lean forward to take a bite.
CR2, who has a C4-C6 Spinal Cord Injury and very limited head and neck mobility, requires inside-mouth bite transfer~\cite{jenamani2024bitetransfer}.
In a social dining scenario, CR1 prefers that the robot retract after bite transfer so they can better see their companion.
In another social scenario, CR2 prefers to control the robot using a custom \textit{long-continuous-open-mouth} gesture that they create themselves through the user interface (a standard open-mouth gesture would be falsely triggered in conversation).
These are a few examples among many underscoring the need for a mealtime assistance system that can be personalized in-the-wild by the users themselves.

To better understand the nature of personalization in mealtime assistance, we start our work with a \emph{formative user study} (Section~\ref{sec:personalization_needs}).
We use speculative videos to guide in-depth conversations with 21 care recipients who have diverse medical conditions and mobility limitations.
This study has two outcomes.
The first is a collection of specific personalization requests that we use to develop our system (Section \ref{subsec:adaptability}).
The second is the realization of three key tenets crucial for personalization in mealtime assistance: \emph{adaptability}, \emph{transparency}, and \emph{safety} (Section~\ref{sec:tenets}).
Care recipients voice the need for systems that can \emph{adapt} to their needs and preferences---not just once, but from meal to meal, and over time as their needs change.
They also express that any adaptations should be \emph{transparent} so that they are able to understand and predict system behavior.
Finally, they underscore the importance of \emph{safety}, especially when the system changes.

With these key tenets in mind, we propose FEAST, a \underline{f}lexible m\underline{e}altime-\underline{a}ssistance \underline{s}ystem \underline{t}owards in-the-wild personalization (Section~\ref{sec:system}).
FEAST is designed for personalization at both the hardware and software level.
The hardware features a modular tool-change apparatus so that a single robot arm can assist with feeding, drinking, and mouth wiping.
The system also features accessible buttons, status LEDs, cameras, microphones, and speakers to enable a wide range of customizable interactions with the user.
We also propose a novel feeding utensil that increases the robot's workspace and decreases obstructions to the user's view.

On the software side, to strike a balance between adaptability, transparency, and safety, we propose to sequence together parameterized behavior-tree-based skills to achieve user-specified goals.
The parameterized behavior trees provide a mechanism for personalization: user requests, formulated in natural language, are translated with a large language model (LLM) into structured updates to the behavior trees, which can then be statically validated for safety.
This structured skill representation can also be analyzed by an LLM in response to user transparency requests.
Finally, using the code synthesis capabilities of LLMs, we enable users to create their own custom head gestures that can be added to the behavior trees and used to interact with the robot.
Users engage with the robot through a flexible web-based user interface.

We develop FEAST using \emph{community-based participatory research}~\cite{hacker2013community} in collaboration with two CRs (Figure~\ref{fig:teaser}). This approach, involving co-design and in-depth evaluation with one or two CRs, is well-established in assistive technology research~\cite{kushalnagar2020teleconference, haidenhofer2024research, fussenegger2022depending} and increasingly common in assistive robotics~\cite{chen2012robots, moharana2019robots, nanavati2023design, padmanabha2024independence, ranganeni2024robots, nanavati2025lessons}. Our collaborations began with introductory video calls in November 2022 and have continued through regular meetings---both virtual and in-person at the CRs' homes. These sessions have been instrumental in gathering feedback on system design, brainstorming studies to identify personalization needs (Section~\ref{sec:personalization_needs}), and piloting developments.

This iterative process led to a five-day in-home evaluation in January 2025 (Section~\ref{sec:evaluation}), where CRs fed themselves six meals across three distinct contexts: personal, watching TV, and social (Figure~\ref{fig:teaser}). Results show that CRs successfully completed meals with few researcher interventions and personalized the system on the fly to their preferences and abilities while reporting low cognitive workload, as indicated by NASA-TLX surveys~\cite{hart1988development}. CRs also rated FEAST highly for real-world applicability, as reflected in Technology Acceptance Model~\cite{davis1989user} survey results. Furthermore, both CRs reported that FEAST provided greater control over their meals and a stronger sense of independence compared to their human caregiver, with one noting that conveying preferences was easier with our system. Finally, to further assess ecological validity, we evaluated FEAST with an Occupational Therapist unfamiliar with the system, who confirms its merits over a no-personalization baseline (Section~\ref{sec:occupational_therapist}).

\definecolor{commercial}{HTML}{ffc09f}
\definecolor{academic}{HTML}{fcf5c7} 
\definecolor{ours}{HTML}{adf7b6}
\begin{table*}[t]
\caption{Comparison between FEAST and other mealtime assistance systems. Adaptability is assessed based on ability to handle open-ended user requests, safety by self-reported adherence to ISO 13482~\cite{iso2014safety}, and transparency by Levels 1-5 of the IEEE Transparency Standard~\cite{ieee2021transparency}.}\label{tab:systems_comparision}
\vspace{-0.2cm}
{\centering

\setlength\extrarowheight{1mm}
\resizebox{\textwidth}{!}{
\begin{tabular}{|ccccccccccccc|} 
\hline
\multirow{2}{*}{\textsc{System}} & \multirow{2}{*}{\textsc{Adaptability}} & \multirow{2}{*}{\textsc{Safety}} & \multirow{2}{*}{\textsc{Transparency}} & \multirow{2}{*}{\textsc{Interface}} & \multicolumn{3}{c}{\textsc{Autonomous Bite Acquisition}} & \multicolumn{2}{c}{\textsc{Autonomous Bite Transfer}} & \multirow{2}{*}{\textsc{\begin{tabular}[c]{@{}c@{}}Minimal\\ Occlusion\end{tabular}}} & \multirow{2}{*}{\textsc{\begin{tabular}[c]{@{}c@{}}Workspace\\ Reachability\end{tabular}}} & \multirow{2}{*}{\textsc{\begin{tabular}[c]{@{}c@{}}Other\\ Tasks\end{tabular}}} \\ 
\cline{6-10}
& & & & & \hspace{0.15cm} \textsc{Seq.} \hspace{0.15cm} & \textsc{Pre-acq.} & \textsc{Acq.} & \hspace{0.2cm} \textsc{Outside} \hspace{0.2cm} & \textsc{Inside} & & &\\
\hline
\rowcolor{commercial} Obi \cite{obi} & - & \checkmark & - & \checkmark & - & - & - & - & - & \checkmark & - & - \\ 
\rowcolor{commercial} Neater Eater \cite{neater} & - & \checkmark & - & \checkmark & - & - & \checkmark & - & - & \checkmark & - & - \\ 
\rowcolor{academic} Park et al. \cite{park2020evaluation} & - & - & - & \checkmark & - & - & \checkmark & \checkmark & \checkmark & - & \checkmark & - \\ 
\rowcolor{academic} Bhattacharjee et al. \cite{bhattacharjee2020moreautonomy} & - & - & - & \checkmark & - & - & \checkmark & - & \checkmark & - & - & - \\ 
\rowcolor{academic} Feel the Bite \cite{jenamani2024bitetransfer} & - & - & - & - & - & - & - & - & \checkmark & - & - & -\\ 
\rowcolor{academic} FLAIR \cite{jenamani2024flair} & - & - & - & - & \checkmark & \checkmark & \checkmark & \checkmark & \checkmark & - & - & -  \\
\rowcolor{academic} REPEAT \cite{ha2024repeat} & - & - & - & - & - & \checkmark & \checkmark & - & - & - & - & -\\ 
\rowcolor{academic} Nanavati et al. \cite{nanavati2025lessons} & - & \checkmark & - & \checkmark & - & - & \checkmark & \checkmark & - & - & - & - \\  
\hline
\rowcolor{academic} \textbf{FEAST (ours)} & \checkmark & \checkmark & \checkmark & \checkmark & \checkmark & \checkmark & \checkmark & \checkmark & \checkmark & \checkmark & \checkmark & \checkmark \\
\hline
\end{tabular}}
\\[3pt]
\tikz\draw[commercial,fill=commercial] (0,0) circle (.8ex); for commercial systems, \tikz\draw[academic,fill=academic] (0,0) circle (.8ex); for academic systems \\
}
\vspace{-0.6cm}
\end{table*}

Overall, our contributions include:
\begin{itemize}
    \item FEAST: A flexible \textbf{mealtime-assistance system} built with \textbf{community-based participatory research}, tackling in-the-wild scenarios by integrating diverse skills, custom tools, a flexible interface, and user personalization.
    \item A user study involving 21 care recipients, which identifies diverse \textbf{personalization needs} for mealtime assistance.
    \item A \textbf{personalization framework} built on three key tenets: (i) adaptability via LLM-based code synthesis, (ii) transparency (Levels 1-5) through LLM summarization, and (iii) safety aligned with ISO/TS 13482 safety principles.
    \item A \textbf{five-day in-the-wild system evaluation} with two community researchers, spanning six realistic meals across three distinct environmental contexts, showcasing FEAST's real-world applicability.
    \item An \textbf{evaluation with an Occupational Therapist} unfamiliar with our system, assessing ecological validity and demonstrating improved performance over a non-personalized baseline.
\end{itemize}

\section{Related Work}

\subsection{Mealtime Assistance}
\label{subsec:related_feeding}

Research on mealtime-assistance systems dates back to the 1970s, with early examples such as the Morewood Spoon Lifter~\citep{vapc1977}. Over the years, several commercial systems have been proposed, such as Winsford Feeder, My Spoon, Neater Eater, Bestic Arm, Meal Buddy, and Obi ~\cite{isira2015survey}. However, their reliance on manually programmed, fixed food acquisition and transfer actions has limited user adoption, and as of now, only Obi and Neater Eater remain commercially available. To address these limitations, recent research focuses on using sensors, such as in-hand cameras, to perceive the environment, and plan and execute autonomous motions.

\textbf{Bite Acquisition and Transfer.}  Recent works have explored autonomous strategies for skewering solid bite-sized foods~\cite{sundaresan2022learning, gordon2020adaptive, gordon2021leveraging, gordon2023towards, feng2019robot}, scooping soft food items~\cite{grannen2022learning, tai2023scone, liu2024imrl}, twirling and grouping noodle-like dishes~\cite{sundaresan2023learning}, and cutting~\cite{zhang2019leveraging}. Building on these works, FLAIR~\cite{jenamani2024flair} introduces a bite acquisition framework consisting of a library of vision-parameterized food manipulation skills that use a fork-based utensil. This framework leverages the commonsense reasoning and few-shot learning capabilities of foundation models to appropriately sequence these skills, enabling the feeding of complete meals while adhering to bite ordering preferences of the user. Various works also propose autonomous bite transfer methods~\cite{gallenberger2019transfer, candeias2018vision, park2020evaluation, belkhale2022balancing, jenamani2024bitetransfer, shaikewitz2022mouth}, which can be broadly categorized into two types. Outside-mouth bite transfer methods~\cite{gallenberger2019transfer,candeias2018vision,belkhale2022balancing} bring food close to the care recipient's mouth, requiring them to lean forward to take a bite. For individuals with severe mobility limitations who cannot lean forward, inside-mouth bite transfer methods~\cite{park2020evaluation,shaikewitz2022mouth,jenamani2024bitetransfer} place food directly inside their mouth. 

Our system builds on FLAIR~\cite{jenamani2024flair} for bite acquisition and Gallenberger et al.~\cite{gallenberger2019transfer} and Feel the Bite~\cite{jenamani2024bitetransfer} for bite transfer, enabling it to feed realistic dishes to users with severe mobility limitations. However, recent studies show that feeding robots often exhibit obtrusive motion~\cite{nanavati2023design, belkhale2022balancing}, limiting their use in social settings, and have restricted workspace~\cite{nanavati2025lessons}, constraining plate and user placement. These limitations hinder personalization, as robots must adapt to social contexts with non-obtrusive motion and allow flexible user and plate positioning. Towards addressing these challenges, and based on feedback from community researchers, FEAST adapts these approaches to a novel feeding utensil (Section~\ref{subsec:hardware}) that improves the robot's workspace and decreases obstruction. 

\textbf{Other Essential Mealtime Tasks.} Several works have explored assistance with drinking~\cite{goldau2019autonomous, schroer2015autonomous}, focusing on grasping a cup and having users drink from the rim. We instead focus on drinking from a cup with a straw, aligning with our community researchers' routines. Unlike prior methods, which assume specific cup colors~\cite{goldau2019autonomous} or pre-grasped cups~\cite{schroer2015autonomous}, FEAST emphasizes real-world flexibility, using an adaptable handle that fits various cups without strict shape or color constraints. Some works also propose methods for mouth wiping~\cite{ikeda2019wiping, hawkins2012informing}. Similar to these, FEAST can pick up a custom wiping tool and position it near the user's mouth, ready for wiping. 


\textbf{Mealtime-Assistance Systems.} Most relevant to our work are autonomous systems that demonstrate integration of various components for feeding a complete meal~\cite{park2020evaluation, gallenberger2019transfer, jenamani2024flair, nanavati2025lessons} (see Table \ref{tab:systems_comparision}). Unlike Park et al.~\cite{park2020evaluation}, which focuses on feeding yogurt, and Bhattacharjee et al.~\cite{gallenberger2019transfer} and Nanavati et al.~\cite{nanavati2025lessons}, which limit bite acquisition to skewering, our system employs a range of skills—skewering, scooping, twirling, grouping, and more—to pick up diverse food items and sequence them over a long horizon for feeding an entire meal. Our system differs from FLAIR~\cite{jenamani2024flair} (which integrates with Feel the Bite~\cite{jenamani2024bitetransfer}) by adapting it to a novel feeding utensil and a web interface for user interaction. More significantly, our system is the first to integrate feeding with other essential mealtime tasks, including drinking and mouth wiping, while also automating user personalization. No prior system has explored feeding in diverse in-the-wild scenarios—except Nanavati et al.~\cite{nanavati2025lessons}, which offers customization of limited system parameters.

\begin{figure*}[!t]
  \centering
    \includegraphics[width=\textwidth]{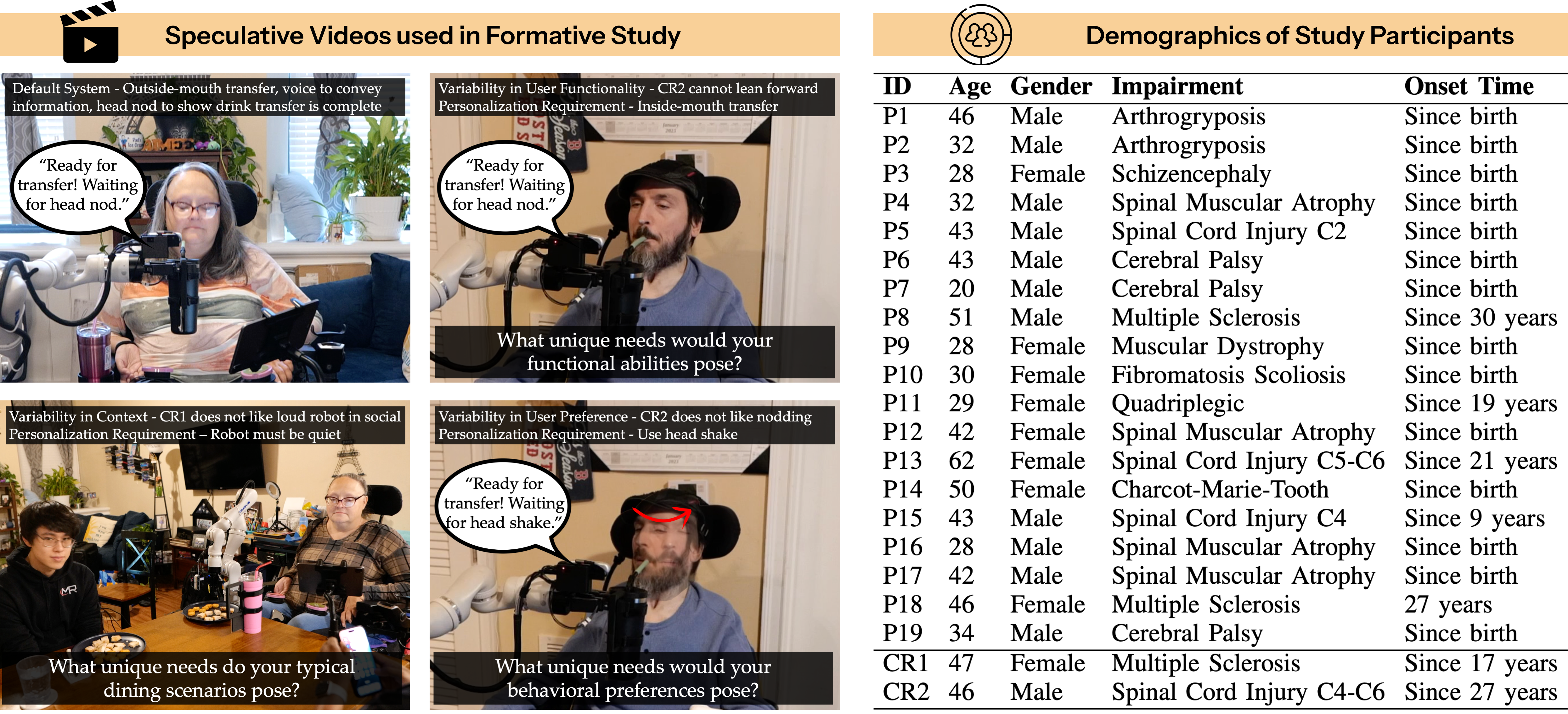}
    \vspace{-0.5cm}
    \caption{We use speculative videos recorded with community researchers (left) to conduct a formative study with 19 care recipients and the 2 community researchers (demographics shown on the right), to identify user personalization requirements. }
    \label{fig:formative_study}
    \vspace{-0.7cm}
\end{figure*}

\subsection{Personalization in Assistive Robotics}

Various formative studies in assistive robotics emphasize the importance of user personalization based on factors such as mobility limitations, behavior, and context~\cite{bhattacharjee2019community, canal2021preferences, nanavati2023design, nanavati2023physically}.

One approach to achieving personalization is through implicit adaptation, where limited test-time examples guide the system to a specific preference~\cite{yang2021desire, he2023learning, liu2024grace, wu2023tidybot, liu2025grace}. FEAST aligns with this by generating personalized gesture detectors using a small number of examples from the user. Unlike prior black-box approaches to gesture personalization~\cite{joshi2017personalizing, brock2021personalization}, FEAST takes a more transparent approach by composing functions from our perception library to generate customized detectors. While some studies evaluate gesture detection for care recipients~\cite{taheri2021designing, jackowski2017head}, FEAST is the first to consider personalized gesture detection for those with mobility limitations.

Another approach involves explicitly identifying preferences through direct user input. Canal et al.~\cite{canal2019adapting} use post-hoc scoring for task planning, but this limits applicability in safety-critical tasks. Nanavati et al.~\cite{nanavati2025lessons} provide predefined customizations for a mealtime-assistance robot through a web interface, while FLAIR~\cite{jenamani2024flair} uniquely explores bite-ordering preferences using language commands. FEAST leverages FLAIR for bite-ordering preferences and significantly extends personalization to other system components by enabling open-ended customization through language inputs.

\subsection{Large Language Models for Human-Robot Interaction}

Recent advancements in large language models (LLMs)~\cite{achiam2023gpt, chowdhery2023palm} enable the use of language as an interface for human-robot interaction. Some approaches utilize LLMs for high-level planning, allowing systems to generate plans over a fixed library of skills based on user language commands~\cite{kambhampati2024can, padmanabha2024voicepilot, chen2023autotamp, huang2022inner, agia2022taps, lin2023text2motion, singh2023progprompt}. Others focus on generating low-level skills by training on large vision-language-action datasets~\cite{brohan2022rt, brohan2023rt, padalkar2023open, kim2024openvla}, synthesizing rewards~\cite{yu2023language, ma2023eureka}, or inferring skills via code generation~\cite{liang2023code, arenas2024prompt}. FEAST represents a recent category of approaches that enable users to edit both high-level plans and parameters of low-level skills using LLMs~\cite{grannen2024vocal, belkhale2024rt, shi2024yell}. However, given the physical nature of human-robot interactions in mealtime assistance, FEAST restricts edits to carefully selected operations to ensure safety.
\section{Identifying Personalization Needs from Care Recipients with Diverse Medical Conditions}
\label{sec:personalization_needs}

\begin{figure*}[!t]
  \centering
    \includegraphics[width=\textwidth]{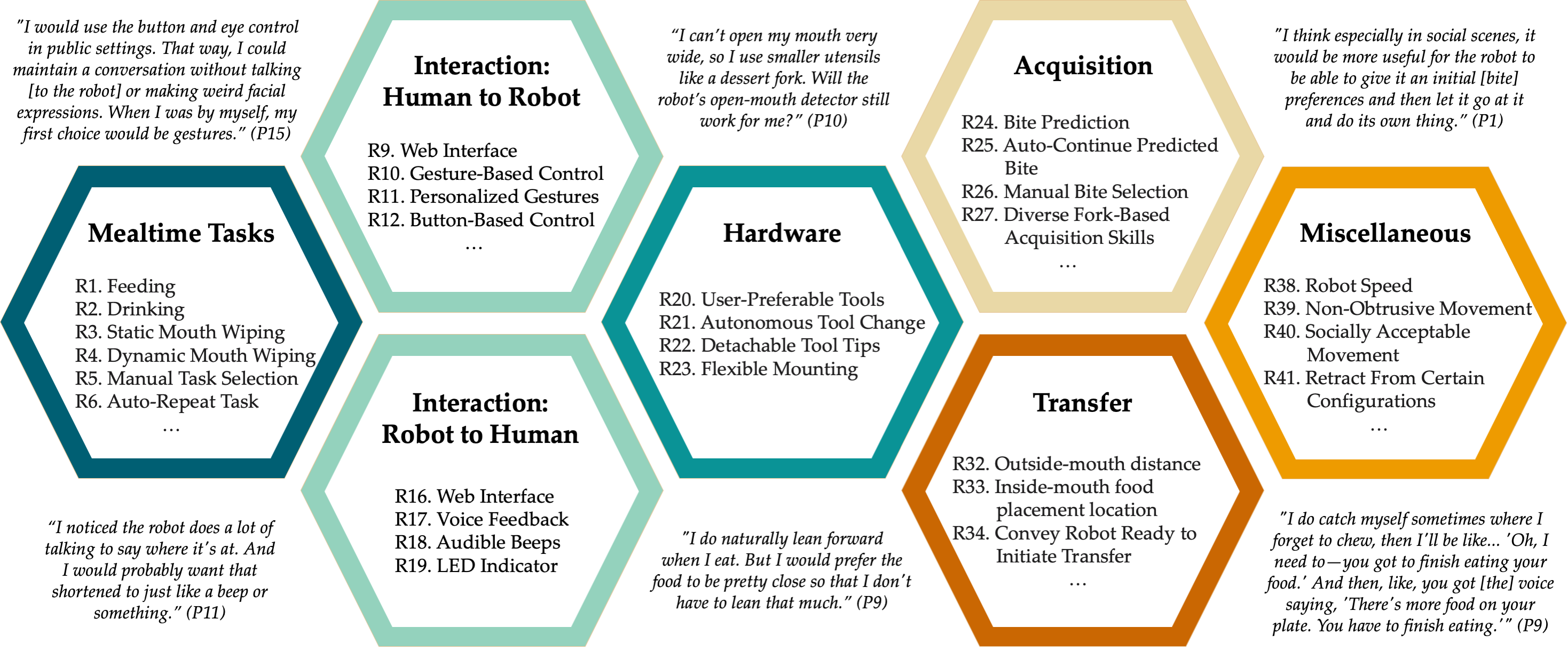}
    \caption{We extract and categorize personalization requests from our formative study with 21 care recipients. See Appendix A for details.}
    \vspace{-0.7cm}
    \label{fig:caregiving_needs}
\end{figure*}

To guide the development of our flexible mealtime-assistance system, we conducted a formative user study with 21 care recipients (including 2 CRs) with diverse mobility limitations. 
Our objectives in this study were two-fold: (1) to collect \emph{specific personalization requests}; and (2) to identify \emph{key tenets} of personalization in the context of mealtime assistance.

\textbf{Participants.}
We recruited 19 care recipients with diverse medical conditions such as Cerebral Palsy, Multiple Sclerosis, Muscular Dystrophy, Spinal Cord Injury, etc. (Figure ~\ref{fig:formative_study}).
There were 8 female and 11 male participants who ranged in age from 20 to 62 years old and who were located throughout the United States.
The participants were also diverse in terms of their prior experience with mealtime-assistance systems: some had no experience; others had participated in academic studies; and some had extensive experience (e.g., P6 noted \textit{``I had a Winsford Self-Feeder for over 20 years until the second one I had stopped working 4 months ago''}).

\textbf{Design Materials.}
Previous work~\cite{madan2022sparcs} has shown that caregiving scenarios vary depending upon (i) care recipient functionality, (ii) care recipient behavior, and (iii) environment.
Working with our community researchers, we identified realistic mealtime assistance scenarios where each of these three components vary.
We then recorded speculative videos of the scenarios using our robot mealtime-assistance system (see Section~\ref{sec:system} for system details).
The videos were organized into a slideshow and supplemented with conversation prompts (Figure \ref{fig:formative_study}).
For example, one video showed the robot successfully feeding a care recipient with outside-mouth bite transfer; the next video showed the robot failing to feed a different care recipient who has very limited neck mobility; a final video in the sequence showed the robot successfully feeding the latter care recipient with inside-mouth bite transfer.
After these three videos, we posed the question: ``What unique needs might your own functional abilities pose for the robot?''
Other speculative videos preceded discussion questions including: ``What unique needs might your own behavioral preferences pose for the robot?'' and ``In what scenarios do you typically eat? What unique needs might these scenarios pose for the robot?''

\textbf{Study Design.}
To broaden participation in the study, we conducted conversations virtually.
Caregivers were also invited to join and actively contribute to the discussion.
Each session consisted of the following steps:
(1) introductions and agenda; 
(2) a brief review of recent progress in robot mealtime-assistance systems and an extension of gratitude to the study participants for facilitating this progress;
(3) speculative videos and discussion questions;
(4) final open-ended discussion and conclusion.
Care recipients were compensated \$30 USD for their participation in the 90-minute study.

\textbf{Outcomes.}
This formative study successfully accomplished its two objectives.
First, we collected specific personalization requests, which we group together into various categories and visualize in Figure \ref{fig:caregiving_needs}. We detail each of these requests in Appendix A. These diverse requests highlight the need for personalization in a mealtime-assistance system and suggest where we should focus our efforts to meet the needs of users.

Our second objective was to identify key tenets for personalization.
From our conversations, we identified three themes: adaptability, transparency, and safety.
\begin{enumerate}
    \item \emph{Adaptability:} The first clear tenet of personalization is adaptability: changing system behavior in response to user requests. Study participants highlighted the need for adaptability beyond one-time system setup. For example, they requested the ability to customize the system depending on the feeding scenario---when watching TV, controlling the robot with a button may be preferable to verbal commands; or when dining socially, the robot should retract to a resting position immediately after bite transfer. Study participants also described how their preferred interaction with the robot may change throughout the day, depending on their energy level, and over time, as their medical conditions change.
    \item \emph{Transparency:} Study participants also emphasized the importance of being able to understand the behavior of the robot, especially if that behavior may change. For example, describing their ideal relationship with the robot, P8 said: \textit{``It would be a symbiotic relationship; I would like to understand the robot capabilities and the robot would understand my requirements.''}
    \item \emph{Safety:} A final key tenet of personalization that emerged from the formative study was safety. For example, P12 said: \textit{``Safety is the most important part. If I turn my head to look somewhere the robot should not try to feed me at that position which might be unsafe.''}
\end{enumerate}

These specific examples and key tenets informed the design of our mealtime-assistance system, which we describe next.


\section{FEAST: A \underline{F}lexible M\underline{e}altime-\underline{A}ssistance \underline{S}ystem
\underline{T}owards In-the-Wild Personalization}
\label{sec:system}

In this section, we present FEAST, a mealtime-assistance system that enables users to personalize to in-the-wild eating scenarios commonly encountered in real-world settings. \emph{All hardware and software components of FEAST are open-sourced on our website.} In the following subsections, we describe our system hardware (Section~\ref{subsec:hardware}), software (Section~\ref{subsec:software}), and user interface (Section~\ref{subsec:interface}), explaining how each component can be personalized while adhering to the tenets of adaptability, transparency, and safety.

\subsection{System Hardware}
\label{subsec:hardware}

\begin{figure*}[!t]
  \centering
    \includegraphics[width=\textwidth]{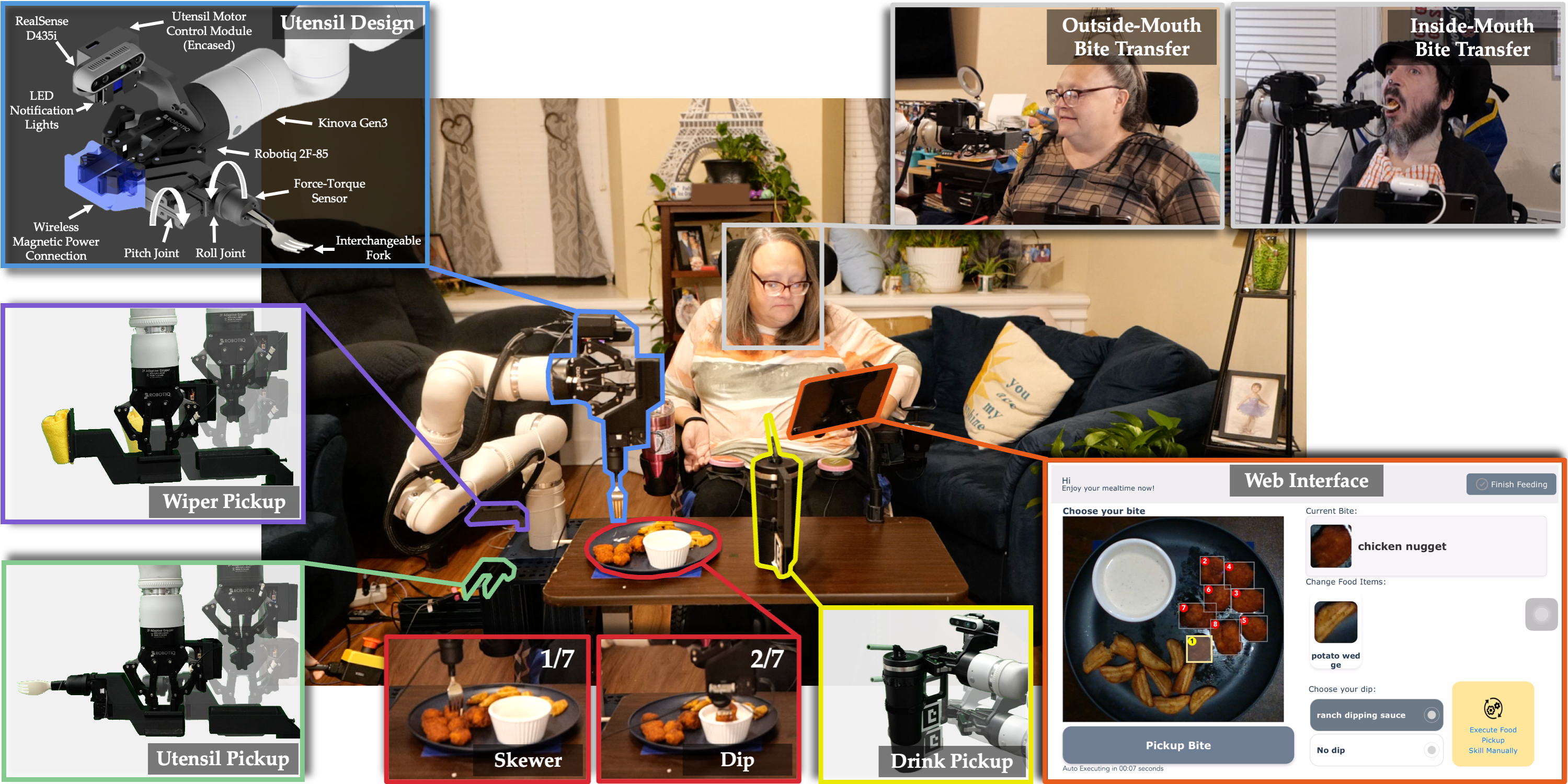}
    \caption{FEAST features diverse mealtime-assistance skills such as feeding, drinking and mouth wiping, custom tools, and a flexible web interface, all of which can be personalized in-the-wild by end users.}
    \vspace{-0.7cm}
    \label{fig:system_details}
\end{figure*}

FEAST (see Figure~\ref{fig:system_details}) uses a Kinova Gen3 7-DoF robot arm~\cite{kinova} and a Robotiq 2F-85 gripper~\cite{robotiq}. It can be flexibly mounted either on the user's ROVI wheelchair~\cite{rovi}, powered by the wheelchair's battery, or on a movable Vention stand~\cite{vention}, powered by a wall outlet. 

\textbf{Tool-Change Apparatus.} FEAST employs three custom tools. First, a novel feeding utensil with integrated motors provides wrist-like degrees of freedom, enabling tasks such as twirling, scooping, and maintaining an upright orientation when holding food. The utensil’s fork, made of compliant silicone, is connected to a 6-axis Nano25 ATI force/torque sensor~\cite{ati}. This default fork is detachable, allowing users to exchange it with a metal fork if desired. To power and control the utensil without dangling wires, the robot’s gripper fingertips are replaced with custom fingertips featuring magnetic electrical connections that engage with complementary connectors on the utensil when grasped. Second, for drinking, FEAST uses an adaptable mug handle inspired by adaptable mug holders~\cite{mugholder}, which accommodates various cup shapes without strict dimensional constraints and features an ArUco marker~\cite{garrido2014automatic} for autonomous grasping. Third, for mouth wiping, a custom tool with a removable microfiber cloth provides gentle cleaning.
Each tool is mounted and dismounted by the robot opening and closing its finger tips (see Figure~\ref{fig:system_details}).

\textbf{Novel Utensil Orientation and Camera Mount.} Previous state-of-the-art feeding systems use forward-facing in-hand cameras and similarly oriented utensils~\cite{jenamani2024flair, jenamani2024bitetransfer, nanavati2025lessons, gallenberger2019transfer}. However, this design restricts the robot's workspace, requires large movements to transition between acquisition and transfer, and can obstruct the user’s view during feeding. Based on feedback from CRs, FEAST introduces a simple and effective solution: positioning tools at a perpendicular angle, mimicking natural human wrist movements during eating. This change, however, requires the camera to align accordingly. To address this, FEAST employs a custom camera mount with an RGB-D Intel RealSense Camera~\cite{camera}, oriented perpendicularly away. 

\textbf{Compute and Networking.} FEAST employs custom controllers, such as a task-space compliant controller operating at approximately 1 kHz. This necessitates a dedicated real-time control system, for which we use an Intel NUC \cite{nuc}. The primary computing platform is a Lenovo Legion Pro 7i laptop equipped with a 16GB RTX 4090 GPU \cite{compute}. Communication between the main compute unit, the NUC, the robot, and the web interface is managed via a Nighthawk RAX43 router \cite{router}. When FEAST is mounted on a wheelchair, these components can also be securely mounted to ensure they move along.

\textbf{Accessible Buttons.} FEAST utilizes three accessible buttons~\cite{estop}. Two are intended for the user: one customizable through our personalization pipeline for interactions, and another serving as an emergency stop. The third button functions as an experimenter emergency stop.

\textbf{Status LED.} FEAST features a status LED within the camera mount to alert users when their attention is needed, based on feedback from CRs. This is especially useful for users multitasking, such as watching TV while eating.


\subsection{System Software}
\label{subsec:software}

FEAST sequences together skills to accomplish goals given by the user.
For example, if the user requests a drink while the robot is holding the utensil, the robot would invoke skills to (1) dismount the utensil; (2) mount the drink; and (3) transfer the drink to the user.
We next describe how skills are generally implemented, adapted, and sequenced together, before detailing the specific skills used in this work.

\textbf{Skills as Parameterized Behavior Trees.}
We implement skills as behavior trees~\cite{colledanchise2018behavior}.
As an extension of standard behavior trees, we expose \emph{node parameters} that can be adapted in response to user requests (see below).
For example, the behavior tree for bite acquisition includes three parameters: \texttt{Speed}, \texttt{TimeToWaitBeforeAutocontinue}, and \texttt{AskUserForConfirmation}.
Every parameter is associated with a domain of possible values.
For example:
\begin{itemize}
    \item $\texttt{Speed} \in \{\texttt{low}, \texttt{medium}, \texttt{high}\}$
    \item $\texttt{TimeToWaitBeforeAutocontinue} \in [5, 100]$
    \item $\texttt{AskUserForConfirmation} \in \{\texttt{true}, \texttt{false}\}.$
\end{itemize}
All nodes and parameters are given human-readable names and descriptions to facilitate LLM-based adaptations.

\textbf{Personalizing Skills from Natural Language.} FEAST users can make personalization requests through spoken or typed natural language.
Our pipeline for processing these requests is as follows:
\begin{enumerate}
    \item The natural language request is converted into structured behavior tree updates using an LLM. (adaptability)
    \item Each potential update is checked for safety. If the updates are deemed safe, the behavior trees are updated. Otherwise, a failure is reported. (safety)
    \item The updates and outcomes are briefly summarized with an LLM and reported back to the user. (transparency)
\end{enumerate}
We now describe these steps in more detail.

\textit{1) Language $\to$ Structured Updates.}
We use an LLM (GPT-4o~\cite{hurst2024gpt}) to translate natural language personalization requests into structured behavior tree updates.
The LLM is prompted with a brief explanation of the scenario, the personalization request, the behavior trees, and the form of an expected response.\footnote{We also found that prompting the LLM to rephrase the original request into a ``specific setting that should be changed in the robot's software'' helped with certain requests such as ``The robot is too slow right now.''}
The output of the LLM is statically checked to ensure that the structured requests are in the expected form and the names of behavior tree nodes and parameters are valid.
If any of these checks fail, the LLM is automatically re-prompted (up to 3 times in experiments) with failure information.
It is important to note that one personalization request can elicit multiple updates.
For example, if the user requests for the robot to ``always move as fast as possible,'' multiple behavior trees with speed parameters would be updated.

\textit{2) Safety Checks.}
If the structured behavior tree requests pass static checks, we pass them through another round of safety checks.
We consider two types of updates: node additions and parameter changes.
In this work, to guarantee safety, we permit only three types of node additions: ``pause'', ``wait for gesture'', and ``retract.''
We further restrict ``retract'' node additions to ensure that the robot only retracts from a limited set of configurations that have been empirically tested.
For parameter changes, we ensure that the proposed values lie within the parameter domains.
These domains are carefully chosen to ensure that safety is guaranteed, while also giving the user enough flexibility to meet their personalization needs.

\textit{3) Update Results $\to$ Language.} Given the history of structured behavior tree updates and the check outcomes, we again use an LLM to generate a brief summary.
The LLM is prompted with a brief explanation of the context, the original user request, the structured updates, and the outcomes, and asked to generate a short summary that a non-technical end user would understand.
This summary is displayed on the user interface (Section~\ref{subsec:interface}).

This process---converting natural language requests into structured behavior tree updates that can be checked for safety and summarized back to the user---addresses our three key tenets of personalization.
By exposing parameters and allowing for certain node additions, the system is \emph{adaptable}; in summarizing the process, the system is \emph{transparent}; and by restricting and checking the adaptations, the system is \emph{safe}.

\textbf{Synthesizing New Gestures with LLMs.}
As an additional step towards open-ended personalization, we allow the user to synthesize new gestures that can then be integrated into the behavior tree---for example, if the user wishes to use a custom head shaking motion to indicate that they are ready for bite transfer.
Through the interface (Section~\ref{subsec:interface}), the user is guided through the process of collecting a few positive and negative examples of their new gesture.
The user also gives the gesture a name (e.g., ``head shake'') and a brief description (e.g., ``shaking my head from right to left'').
FEAST then uses LLM-based program synthesis to generate a programmatic gesture detector that takes a head tracking module as input.
Our program synthesis pipeline is also able to propose and optimize hyperparameters, e.g., \texttt{NumberHeadShakes} or \texttt{NoiseTolerance}. See Appendix B for details.

\smallskip
\textbf{Sequencing Skills with PDDL.} Towards generating a sequence of skills to achieve a user's goal, we associate each skill with a PDDL operator~\cite{mcdermott2000pddl}.
For example:
{\footnotesize
\begin{verbatim}
(:action PickTool
  :parameters (?tool - tool)
  :precondition (and (GripperFree) (Reachable ?tool))
  :effect (and (Holding ?tool) (not (GripperFree))
               (not (Reachable ?tool)))
)
\end{verbatim}
}
\noindent where $\texttt{?tool}$ can be grounded with \texttt{utensil}, \texttt{mug}, or \texttt{wiper}.
See McDermott et al.~\cite{mcdermott2000pddl} for a formal introduction to PDDL.
We use an optimal task planner (FastDownward~\cite{fd} with alias $\texttt{seq-opt-lmcut}$) to sequence the skills together given a known initial propositional state and a propositional goal derived from the user's input.
For example, when the user requests a bite, FastDownward generates a three-step plan:
$\texttt{PickTool(utensil)}$,
$\texttt{AcquireBite(utensil)}$, 
$\texttt{TransferTool(utensil)}$.
The plan is executed open-loop and the propositional state is updated using the operator effects.
If the user subsequently requests a drink after requesting a bite, the second plan would be:
$\texttt{PlaceTool(utensil)}$,
$\texttt{PickTool(mug)}$, 
$\texttt{TransferTool(mug)}$.

\textbf{Skills Included in FEAST.} We now detail the specific skills that are currently implemented in FEAST.
\begin{itemize}[leftmargin=*, labelindent=0pt]
    \item \texttt{PickTool}: Grasping routine for \textit{utensil}, \textit{mug}, or \textit{wiper}. The utensil and wiper are placed in a fixed tool mount relative to the robot's base, enabling a predefined pickup motion. For the mug, the robot detects the ArUco marker on the adaptable handle and executes an appropriate grasp.
    \item \texttt{PlaceTool}: Returns the tool to its last known location. For utensils or wipers, this is a predefined position on the tool mount; for the mug, it is the original pickup position.
    \item \texttt{AcquireBite}: Picks up food from the plate. FEAST builds on the bite acquisition framework proposed by FLAIR~\cite{jenamani2024flair}, adapting it to our novel utensil. This framework includes a vision-parameterized food manipulation skill library with four pickup actions---skewering, scooping, twirling, and dipping---and three pre-acquisition actions---pushing, grouping, and cutting---to prepare for future pickups. We incorporate force thresholds via the force-torque sensor to ensure safe operation. For each new bite, FEAST processes the plate image using vision-language foundation models to identify plate contents and predict the required skill sequence for each food type. These predictions, along with the user's bite order preferences and bite history, are provided to an LLM, which determines the next bite while balancing preference adherence and efficiency. The robot then executes the predicted skills to pickup the bite. 
    \item \texttt{TransferTool}: By default, FEAST performs outside-mouth transfer~\cite{gallenberger2019transfer}, where it detects the user's head pose once and moves to a predefined distance outside the mouth. Upon user request, FEAST can switch to inside-mouth transfer~\cite{jenamani2024bitetransfer}, using a task-space compliant controller to continuously track the user's head pose and bring the tool inside the mouth. After detecting that the user has finished taking a bite, sip, or completed mouth wiping, it returns to its initial position. Both transfer methods utilize the head perception pipeline proposed by Feel the Bite~\cite{jenamani2024bitetransfer}.
    \item \texttt{EmulateTransfer}: Moves the gripper to a configuration just in-front of the user's mouth. Useful for recording and testing new gestures as we detail in the next section. 
    \item \texttt{Retract}: Moves the robot to the retract configuration with an empty gripper.
\end{itemize}

\subsection{User Interface}
\label{subsec:interface}

FEAST features a web-based interface accessible on personal devices, such as tablets or phones, enabling integration with assistive technologies that care recipients already use. For example, CR2 uses a tracker that follows a reflective dot on their nose~\cite{nosetracker} (Figure \ref{fig:system_details}). We refine the interface design through multiple iterations with CRs to address their needs; they highlight that the interface should have big buttons that are easy to click, and should support speech to text. 

We implement the interface using Vue.js~\cite{vue}, an open-source front-end JavaScript framework, while ROS Noetic~\cite{ros_noetic} facilitates communication with the mealtime-assistance skills. FEAST’s web interface displays pages on demand from the robot, automatically updating to reflect system behavior as users personalize it to meet their evolving needs. The interface supports system adaptability and transparency via a personalization webpage, as we detail in Section ~\ref{subsec:procedure}.

Beyond fully autonomous workflows that can continuously feed users without intervention, FEAST’s interface allows on-demand user control. Users can override the robot’s autonomous predictions, for instance, by manually selecting the next bite or pinpointing key points on a displayed plate image as parameters to bite acquisition skills. Such overrides help users recover from autonomous errors, increasing the robustness of the system in in-the-wild scenarios~\cite{hawes2017strands, wang2018tritonbot}.

\subsection{Default Mealtime-Assistance Procedure}
\label{subsec:procedure}

\textbf{Mealtime Assistance.} At the start of a new meal, the robot begins in the ``retract configuration" without holding any tool. The user is presented with the \emph{New Meal Page} on the web interface, where they specify the food items on the plate and their preferred bite order (e.g., ``Feed me all of X first, then Y"). The interface then transitions to the \emph{Task Selection Page}, allowing the user to choose from three primary actions: requesting a bite, taking a sip, or wiping their mouth, which sends $\texttt{TransferTool(utensil)}$, $\texttt{TransferTool(mug)}$, and $\texttt{TransferTool(wipe)}$ commands respectively. If the robot is not holding the appropriate tool, the task planner directs it to place the current tool and pick up the correct one. Moreover, for a bite, the task planner invokes $\mathrm{AcquireBite}$, which looks at the plate contents and predicts the next bite in adherence to the user's ordering preferences. The interface subsequently opens a \emph{Bite Acquisition Page}, where the user can:
\begin{enumerate}[leftmargin=*, labelindent=0pt]
    \item Switch to a different bite of the same or another food type.
    \item Switch to a \emph{Manual Bite Acquisition Page} to choose a specific skill (e.g., skewering) and set keypoint parameters (e.g., the exact skewering point on the plate image).
\end{enumerate}
If the user clicks ``acquire bite" or does not press any button within 10 seconds, the robot proceeds with bite acquisition. For $\texttt{TransferTool}$, the interface prompts the user to confirm readiness for transfer. Once confirmed, the robot moves to a ready-to-transfer configuration and announces, ``Please open your mouth when ready.'' It waits for the user to open their mouth and then, moves the fork, straw, or wiping tool tip near the user's mouth, saying, ``Ready for transfer.'' After detecting a bite via the force/torque sensor, or a completed sip or wipe through a head nod, the robot retracts to the ready-to-transfer position. It then returns to the Next Task Selection Page, with a 10-second auto-continue countdown to repeat the last task if it was a bite or sip. At any point during the meal, the user can click a ``finish eating" button that triggers the $\mathrm{Retract}$ skill. 

\textbf{Personalization.} Before and during a meal, users can access the \emph{Personalization Page}, where they can send:
\begin{enumerate}[leftmargin=*, labelindent=0pt]
    \item Adaptability Requests: Users can input commands to personalize components of the mealtime-assistance system to meet their needs (details in Section~\ref{subsec:adaptability}).
    \item Transparency Requests: Users can ask about the robot's behavior, and the system responds through textual answers (details in Section~\ref{subsec:transparency}).
\end{enumerate}

\begin{figure*}[!t]
  \centering
    \includegraphics[width=\textwidth]{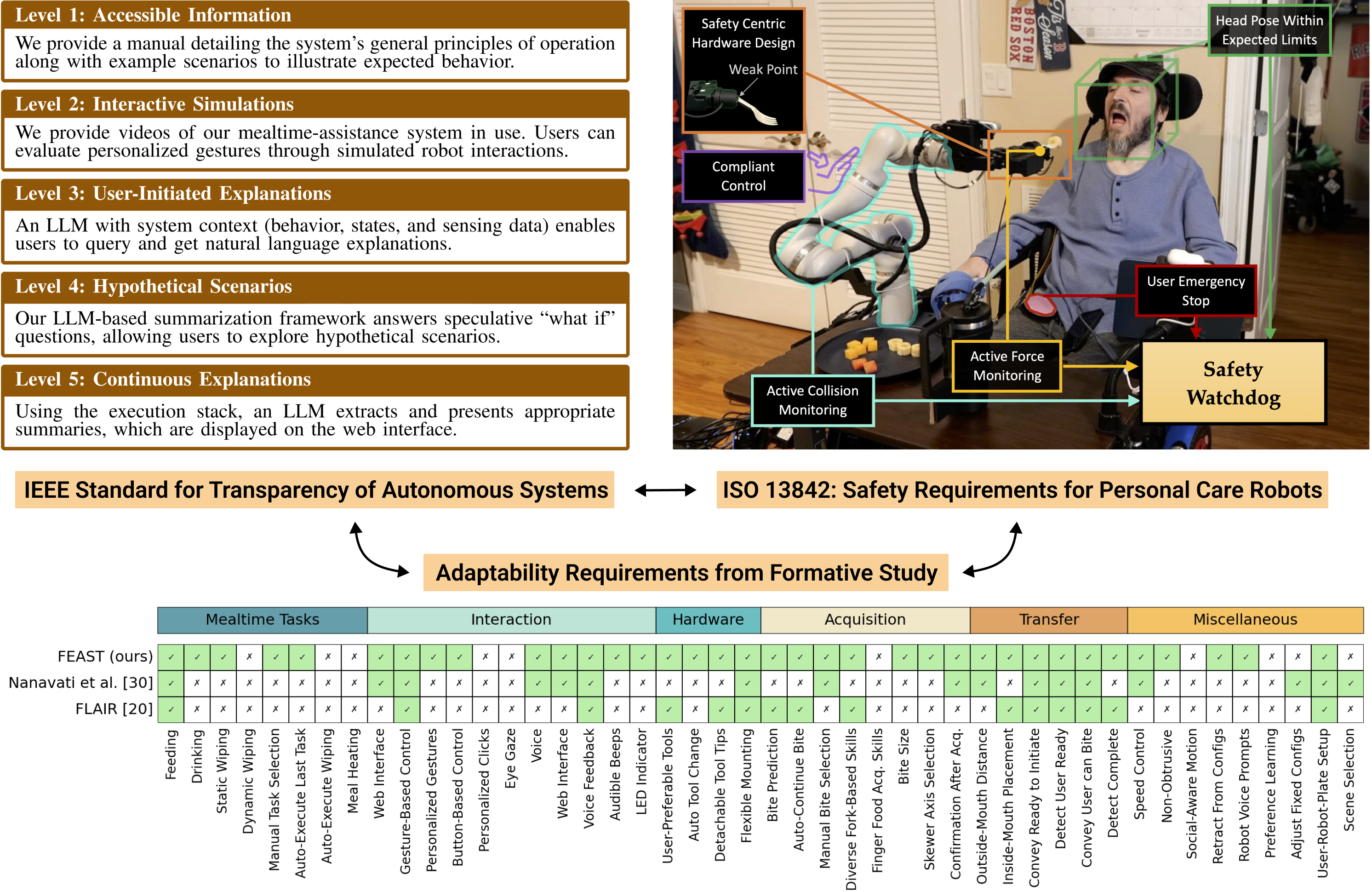}
    \vspace{-0.5cm}
    \caption{FEAST proposes a personalization framework for mealtime assistance built on three key tenets: (i) adaptability towards tackling the various request types identified through our formative study, (ii) transparency towards Levels 1-5 on the IEEE Transparency Standard, and (iii) safety aligned with ISO/TS 13482 safety principles. }
    \label{fig:personalization_tenets}
    \vspace{-0.7cm}
\end{figure*}

\noindent The page also allows users to transition to the \emph{Gestures Page}, which focuses on adding personalized gestures to the system's library. To add a new gesture, users provide a language label and description. The interface then triggers $\texttt{EmulateTransfer}$, moving the robot to a position just outside the user’s mouth without holding a utensil to record positive and negative gesture examples. Once the recordings are complete, the robot returns to the ready-to-transfer configuration, and the interface displays the accuracy of the newly trained gesture detector. Users can also test gesture detectors by having the robot move to the position just outside their mouth (without holding a utensil) while the interface displays whether the detector is triggered by their movements. When switching between webpages, the interface displays a white screen with dynamic text, continuously explaining its current actions. Together, these personalization features enhance adaptability and align with the IEEE Transparency framework (Section~\ref{subsec:transparency}).


\section{FEAST Achieves the Three Tenets of Personalization}
\label{sec:tenets}

Our formative study affirmed the importance of adaptability, transparency, and safety for a mealtime assistance system that is personalized in-the-wild.
In this section, we detail how the system design of FEAST addresses these three tenets.

\subsection{Adaptability}
\label{subsec:adaptability}

In our formative user study, care recipients voiced 46 types of adaptability requests across 7 categories (Figure~\ref{fig:caregiving_needs}). Requests of the same type often differed from user to user. For example, P1 said: \textit{``I would want the robot to be fast, as fast as it could go maybe up until those very last moments of moving towards my head,''} while P9 said, \textit{``I have a trach and have to eat slowly.''} P10 said: \textit{``It would be useful to give the robot more autonomy during those social situations, to pick the bite or whatever, rather than me specifying something exactly,''} and P5 said, \textit{``This might be just a personal thing, but the idea of hitting a switch and then waiting for something to happen makes me then feel like I'm no longer in control.''} Users also voiced the importance of adapting over time: for example, P8 mentioned, \textit{``I'm able to do that right now. But these things might change over time. So you might want to keep updating the robot with the progression of the disease.''}

In Figure~\ref{fig:personalization_tenets}, we compare FEAST to two state-of-the-art feeding systems, Nanavati et al.~\cite{nanavati2025lessons} and FLAIR~\cite{jenamani2024flair}, based on their ability to address the 46 adaptability request types identified in our study. FEAST covers 36, compared to 17 for Nanavati et al. and 15 for FLAIR. Appendix A provides detailed justifications for each system. Appendix B presents an ablation study demonstrating that our LLM-based personalized gesture synthesis outperforms a no-personalization baseline.

\subsection{Transparency}
\label{subsec:transparency}

The IEEE Standard for Transparency of Autonomous Systems~\cite{ieee2021transparency, winfield2021ieee}, which explicitly includes care robots within its scope, defines five levels of transparency requirements, ranging from 1 to 5 (Figure~\ref{fig:personalization_tenets}). FEAST aims to meet all these levels.

Level 1 specifies that users must have access to information such as example scenarios and general principles of operation. FEAST satisfies this requirement with an instruction manual provided to new users. Level 2 requires interactive training material, enabling users to rehearse their interactions with the system in virtual scenarios. To address this, we provide demonstration videos of the system across various use cases. Moreover, after adding personalized gestures, users can test them directly on the real robot in isolation.

Levels 3 and 4 focus on providing user-initiated explanations. Level 3 requires explanations of the system's most recent activity, while Level 4 demands answers to “what if” queries about system behavior. FEAST uses LLM summarization to fulfill these requirements. We define the following to encapsulate the current system state:
\begin{enumerate}[leftmargin=*, labelindent=0pt]
    \item Current Behavior: Current behavior tree encodings for mealtime-assistance skills, with text descriptions for each node and its parameters.
    \item Node Execution History: A real-time log of behavior tree nodes in-execution and completed.
    \item Perception Log: A real-time log of perceived data, such as plate contents, drink pose, user head pose, and gestures.
    \item Safety Log: A real-time log of safety checks, including any invalidated specifications.
\end{enumerate}
On the Personalization Page of the web interface, users can submit language queries. The LLM processes these queries--along with the current system state, detailed descriptions of system operations (as outlined in Section~\ref{sec:system}), and prior user queries and the corresponding LLM’s responses--to generate concise, understandable explanations for non-expert users. 

Level 5 transparency requires the system to provide continuous explanations of its behavior. FEAST aims to achieve this using LLM summarization. At fixed-time intervals, the system checks for changes in the system state as defined above. If a change is detected, the LLM is queried with detailed descriptions of system operations, the previous state, and the current state, prompting it to generate an explanation of what changed during that period. This explanation is displayed on the web interface whenever no other page is active.



\subsection{Safety}
\label{subsec:safety}

FEAST fits within the scope of ISO 13842: Robots and Robotic Devices--Safety Requirements for Personal Care Robots~\cite{iso2014safety}, categorized as a ``physical assistant robot", with the following hardware and software checks (Figure~\ref{fig:personalization_tenets}):

\noindent \textbf{Pose filtering.} Our pose detection methods (food, drink, and head pose) use predefined zones based on expected plate, drink, and head positions, initialized at the start of each mealtime assistance scenario. Head pose limits are tailored to the user’s range of motion~\cite{reese2016joint}. If detections occur outside these zones---such as mistakenly identifying a caregiver's head pose behind the user during feeding (a scenario that can arise in real-world settings)---the robot transitions to a safety state.

\noindent \textbf{Active collision monitoring.} While our robot verifies its motion plans for collisions with known obstacles, it also uses active collision monitoring to handle unexpected collisions when not using compliant control. This is achieved by comparing torque feedback from the joint F/T sensors with torque predictions from a nominal model-based control law~\cite{haddadin2017robot}.

\noindent \textbf{Compliant control.} For inside-mouth transfer, the most physical human-robot interaction intensive aspect of mealtime assistance, our robot switches to a custom Cartesian-space compliant controller that leverages Damped Least Squares~\cite{nakamura1986inverse}.

\noindent \textbf{Safety-centric hardware design.} Within the feeding utensil, the fork is attached to a custom-designed holder with a mechanical weak point that yields under excessive force, breaking the utensil into two pieces. This safety feature ensures that if excessive force is applied while the utensil is in the user’s mouth, only the fork tip remains, halting all physical interaction with the robot. For safety, we restrict outside-mouth bite transfers to the default silicone fork. Similarly, the drinking utensil uses a silicone straw, and the mouth wiper features a soft tip.

\noindent \textbf{Emergency Stops.} At all times, both the user and an experimenter have access to physical emergency stop buttons, which immediately transition the robot into a safety state.

\noindent \textbf{Watchdog.} A watchdog continuously monitors the functionality of all robot sensors and emergency stops, transitioning the robot to a safety state if an issue is detected.

\noindent We provide implementation details for the above and further discuss alignment with ISO 13842 guidelines in Appendix C.

\section{Five-Day In-the-Wild Evaluation}
\label{sec:evaluation}

\begin{figure*}[!t]
  \centering
    \includegraphics[width=\textwidth]{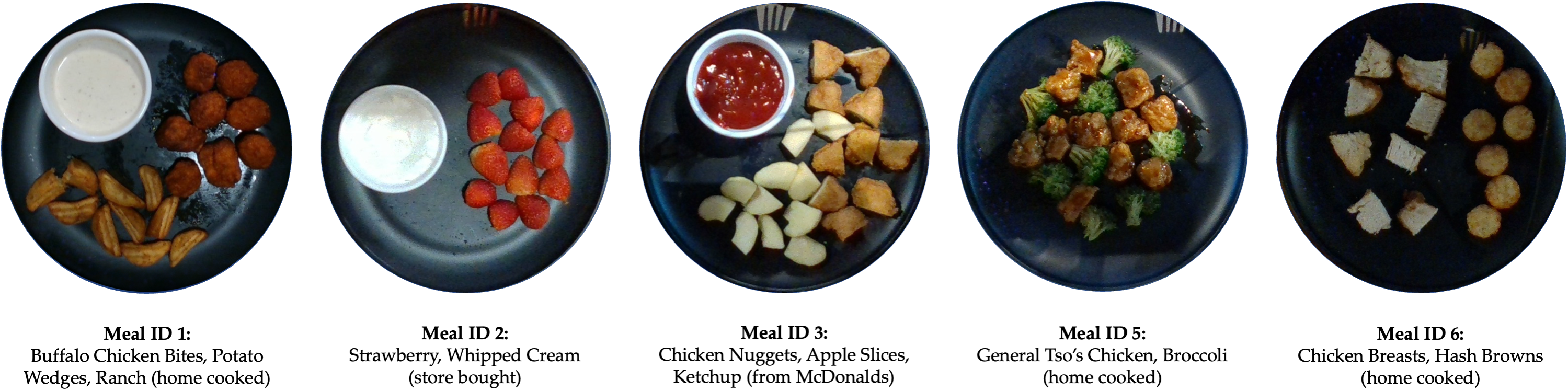}
    \caption{Meals fed to care recipients. The images, captured by the robot's in-hand camera, highlight the variability in lighting conditions. Each meal also included water. Meal ID 4 consisted only of a protein shake, which CR2 usually has for lunch.}
    \label{fig:study_plates}
    \vspace{-0.6cm}
\end{figure*}

To showcase the real-world applicability of FEAST, we evaluate our system with two care recipients (CR1 and CR2) in a five-day in-home study spanning six realistic meals across three distinct environmental contexts:
\begin{enumerate}
    \item \textbf{Personal context} where they focus solely on eating.
    \item \textbf{Watching television} while they are eating.
    \item \textbf{Social context} where they eat with another researcher.
\end{enumerate}
\noindent 

This study asks the question: \textbf{can end-users effectively personalize FEAST to meet their requirements across various mealtime scenarios?} 

\subsection{Study Procedure}

\textbf{Introductory Meals and System Familiarization.} Before any evaluation meals, we conduct one introductory meal with each care recipient on Day 1 of the study to familiarize them with the system. During these training meals, we explain the components of the mealtime-assistance system, demonstrate its various personalization features, and answer their questions.

\textbf{Evaluation Meals.} Over the course of the next 4 days, we feed 3 meals each to the two CRs, in their own homes, in the following chronological order:
\begin{itemize}
    \item \textbf{Meal ID 1} (on Day 2): CR1 Dinner in Personal Context – buffalo chicken bites, potato wedges, and ranch dressing (home cooked meal), served with water.
    \item \textbf{Meal ID 2} (on Day 2): CR1 After-dinner dessert while Watching TV - strawberries with whipped cream (store-bought meal), served with water.
    \item \textbf{Meal ID 3} (on Day 3): CR2 Dinner in Personal Context - chicken nuggets, apple slices, and ketchup purchased from McDonald's, served with water.
    \item \textbf{Meal ID 4} (on Day 4): CR2 Lunch in Social Context - protein shake (CR2's usual lunch). Mug is cleaned and refilled with water after the shake is finished upon request.
    \item \textbf{Meal ID 5} (on Day 5): CR2 Dinner while Watching TV - General Tso's Chicken and Broccoli (home cooked meal), served with water.
    \item \textbf{Meal ID 6} (on Day 5): CR1 Dinner in Social Context - chicken breast strips and hash browns (home cooked meal), served with water.
\end{itemize}
We select the meals for the study based on the care recipient's eating habits, the robot's capabilities, and the need for variation. While we validate our system with similar meal types during study preparation, the exact meals naturally vary based on how they are prepared and what is available. Many adjustments were made on the fly; for example, hash browns from CR1's fridge were added to Meal ID 6 at the last minute.

\textbf{Environmental and User-Specific Variability.} Care recipients live in different homes and have unique preferences for their environment setup. This leads to variations both across and within the same context (see Figure \ref{fig:teaser}), including:

\begin{itemize}
    \item \textbf{Tools:} CR1 uses a metal fork for all meals and a pink mug for the last meal. CR2 uses a silicone fork (safer for inside-mouth transfer) and a black mug for all meals.
    \item \textbf{Web Interface Interaction:} Both use iPads. CR1, with limited limb mobility, interacts using their left arm, while CR2, with very limited limb mobility, uses a mouse tracker that follows a reflective dot on their nose.
    \item \textbf{Eating Tables:} Different table types include multiple overbed tables, a coffee table, and a dining table.
    \item \textbf{Seating Arrangements:} While watching TV, both CR1 and CR2 face the screen. In a social setting, CR1's social partner sits to their right, whereas CR2's social partner sits directly in front.
\end{itemize}

\textbf{Meal Setup and Scene Configuration.} For each new scene, experimenters update scene parameters used for safety checks, such as table height (ensuring a lower limit for acquisition actions) and the user's head pose (used for filtering head perception). They also adjust the fixed configurations that the robot moves to before detecting the plate, drink, and user’s head to ensure their visibility. After the CRs personalize the robot for their personal contexts in their first evaluation meals, we initialize their other contexts with those settings.

\subsection{Evaluation Metrics}

Before the study, participants rate their experiences with human caregivers during mealtime on a 5-point Likert scale, which is later compared to FEAST. To evaluate system performance, we log experimenter interventions for each meal, categorized into hardware, web interface, skills, and personalization. We also track experimenter explanations, recording the number of questions participants ask. Skill success rates are measured for autonomous actions, which rely on perception, such as bite acquisition, bite transfer, drink pickup, drink transfer, and wipe transfer. Fixed-motion actions, such as utensil and mouth wipe pickup from the tool mount and subsequent stow, always succeed. After each meal, participants complete a NASA-TLX~\cite{hart1988development} survey (7-point Likert scale) to assess cognitive workload. At the end of the study, a Technology Acceptance Model (TAM)~\cite{davis1989user} survey evaluates overall satisfaction and usability, along with additional questions assessing key personalization tenets: adaptability, transparency, and safety. Full questionnaires are in Appendix D.

\begin{table*}[ht!]
\caption{Per-meal system performance, including experimenter interventions, explanations, and skill success rates.}
\label{table:in_the_wild_performance}
{\centering
\setlength\extrarowheight{1mm}
\resizebox{\textwidth}{!}{
\begin{tabular}{lcccccccccc} 
\hline
\multirow{2}{*}{\textsc{Meal}} & \multicolumn{4}{c}{\textsc{Interventions}} & \multirow{2}{*}{\textsc{Explanations}} & \multicolumn{5}{c}{\textsc{Skill Success Rate}} \\ 
\cline{2-5} 
\cline{7-11}
& \textsc{Hardware} & \textsc{Web Interface} & \textsc{Skill} & \textsc{Personalization} & & \textsc{Bite Acq.} & \textsc{Bite Transfer} & \textsc{Drink Acq.} & \textsc{Drink Transfer} & \textsc{Wipe Transfer}\\
\hline
\#1: CR1 Personal & 4 & 2 & 1 & 0 & 8 & 24/25 & 14/17 & 3/7 & 2/3 & 1/1 \\
\#2: CR1 TC & 1 & 1 & 0 & 0 & 1 & 22/27 & 11/11 & 1/1 & 1/1 & -\\
\#3: CR2 Personal & 1 & 1 & 0 & 0 & 6 & 29/37 & 15/17 & 3/3 & 3/3 & 1/1\\
\#4: CR2 Social & 1 & 0 & 1 & 0 & 2 & - & - & 2/2 & 5/5 & - \\
\#5: CR2 TV & 0 & 4 & 0 & 0 & 1 & 19/21 & 18/19 & 4/4 & 2/2 & 2/2\\
\#6: CR1 Social & 0 & 0 & 0 & 1 & 2 & 12/12 & 12/12 & 2/2 & 2/2 & 1/1\\
\hline
\end{tabular}}}

\end{table*}

\begin{figure*}[!t]
  \centering
    \includegraphics[width=\textwidth]{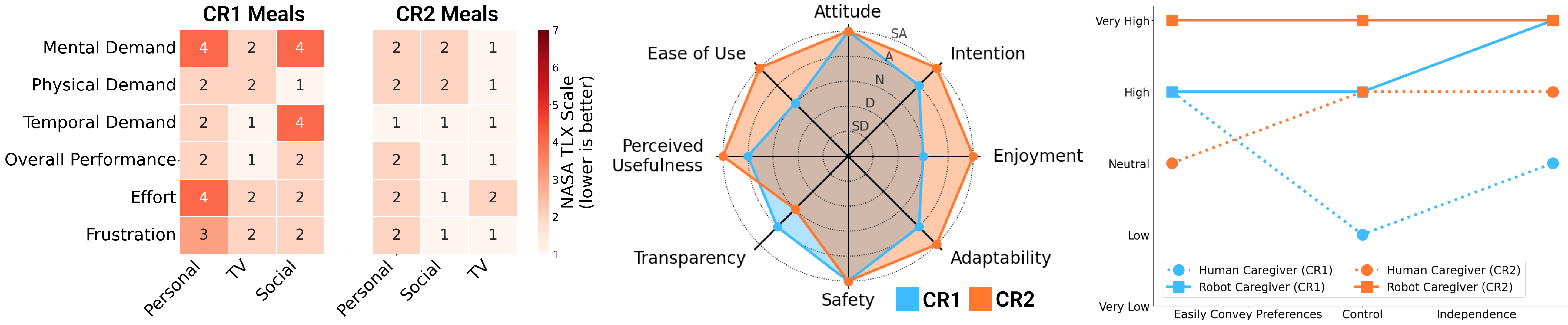}
    \caption{Left: Per-meal NASA-TLX survey results, with CR1 and CR2 meals shown in chronological order. Center: TAM results, including questions on personalization tenets. Right: Comparison between human caregivers and FEAST.}
    \label{fig:study_results}
    \vspace{-0.5cm}
\end{figure*}

\subsection{Results}

\underline{For each meal, CR1 and CR2 reported either fully clearing} \underline{their plate or eating until they were full.} Table \ref{table:in_the_wild_performance} summarizes objective system performance per meal, whereas Figure \ref{fig:study_results} shows subjective metrics. Details on each of the interventions, explanations, and personalization requests made during the study can be found in Appendix E.

\textbf{Interventions.} A total of 18 experimenter interventions occurred across six meals (\(\mathbf{3 \pm 2}\)). Four hardware interventions in the first meal were due to a damaged ArUco marker on the mug handle, which was later replaced. All but one web interface intervention resulted from a recurring network issue that intermittently blocked commands, requiring a refresh. 

\textbf{Explanations.} Users asked experimenters 20 questions across six meals (\(\mathbf{3.33 \pm 2.69}\)). Many could have been directed to the robot, but the transparency page was only accessible from the task selection page, underscoring the need for transparency throughout the eating process.

\textbf{Skill Success Rates.} FEAST demonstrates robustness to different food items, with a bite acquisition success rate of \(\mathbf{89.27 \pm 9.24}\)\% and a bite transfer success rate of \(\mathbf{93.07 \pm 7.70}\)\% . Prior work suggests that an 80\% acquisition rate is sufficient for use~\cite{bhattacharjee2020moreautonomy}. While drink acquisition success was low for Meal ID 1 due to a damaged marker on the drink handle during transit, it improved significantly after replacement, averaging \(\mathbf{100.0 \pm 0.0}\)\% across Meal IDs 2-5. Drink and wipe transfers were also highly successful at \(\mathbf{94.44 ±\pm 13.61}\)\% and \(\mathbf{100.0 \pm 0.0}\)\% respectively.

\textbf{Cognitive Workload Per Meal.}  The average workload imposed by FEAST was relatively low. On a 0--100 NASA-TLX scale (where lower scores indicate lower workload), CR1 had a mean of \(\mathbf{22.22 \pm 10.02}\),  while CR2 had a mean of \(\mathbf{7.41 \pm 5.79}\), compared to a baseline of 37 \cite{hertzum2021reference,nanavati2023design}. Both care recipients rated their success in achieving mealtime goals very positively: the inverted Performance metric (0--100, lower = better) is \(\mathbf{11.11 \pm 9.62}\) for CR1 and \(\mathbf{5.56 \pm 9.62}\) for CR2. 

\textbf{Comparison with Human Caregiver.} Both CRs rated FEAST as providing greater control over their meals and a greater sense of independence compared to their professional caregivers. CR2 also reported less effort in conveying preferences to the robot than to their human caregiver.

\textbf{TAMs and Personalization Tenets.} Results from the TAM survey indicate that both CRs rated FEAST highly across all metrics, with a mean score of $\geq4/5$ for Perceived Usefulness, Ease of Use, Attitude, Intention, and Enjoyment. They also rated the system highly for adaptability and safety but noted that transparency could be improved. This may be because the transparency page was only accessible from the task selection page on the web interface, which was not always available, requiring them to ask experimenters for explanations.

\subsection{Lessons Learned}
\vspace{0.1cm}
\begin{tcolorbox}[colback=itwasallyellow, colframe=darkyellow, title={\textbf{Lesson 1}}, 
fonttitle=\bfseries, 
boxsep=0.5mm, left=1mm, right=1mm, top=0.5mm, bottom=0.5mm, arc=0mm]
\linespread{0.9}\selectfont
Significant variability exists across in-home eating scenarios, and in-the-wild personalization allows users to uniquely adapt the system to these variations.
\end{tcolorbox}

Across different meals, both CR1 and CR2 made a range of personalization requests. These included adjusting robot speed (\textit{“Feed me as fast as you can,”} Meal ID 1), modifying skill parameters (\textit{“Dip the strawberry deeper into the whipped cream,”} Meal ID 2), changing web interface workflows \textit{(“Don’t show continue pages on the web interface,”} Meal ID 3), and customizing interactions between the robot and user, such as muting the robot (\textit{“Be quiet and do not talk at all,”} Meal ID 5) or changing how transfers are confirmed (\textit{“Use the button to complete a transfer only when taking a sip,”} Meal ID 1).

Even under similar contexts, the CRs had different personalization needs. For example, CR2 added a \textit{“continuous mouth open”} personalized gesture detector for initiating transfers in a social setting, to differentiate from when they are talking. On the other hand, CR1 realized the default mouth-open detector required opening their mouth very wide. This design allowed CR1 to talk without inadvertently triggering the robot, and then open their mouth wide when ready to receive a bite. Thus, CR1 continued using it in social settings.

The spatial arrangement of people and objects also influenced personalization preferences. In CR2’s social context, the social partner sat across the table, so CR2 rarely felt obstructed by the robot: \textit{“I completely forgot about the robot sometimes and had to remember that I had to take a drink.”} Conversely, in CR1’s social context, the social partner was seated to the right—directly across from the robot—prompting CR1 to request the robot retract after every bite (\textit{“Move to retract position after every bite”}) to avoid interference.

\begin{tcolorbox}[colback=itwasallyellow, colframe=darkyellow, title={\textbf{Lesson 2}}, 
fonttitle=\bfseries, 
boxsep=0.5mm, left=1mm, right=1mm, top=0.5mm, bottom=0.5mm, arc=0mm]
\linespread{0.9}\selectfont
Transparency helps users iteratively refine the system to meet their preferences, even when adaptability commands are not always effective.
\end{tcolorbox}

FEAST's adaptability leverages an LLM to process language commands and update the robot's behavior. However, LLMs can sometimes hallucinate or make mistakes. For example, when CR1 tried to switch to a button for wipe transfers with the command, \textit{``Use button when completing a transfer when taking a sip,''} the LLM mistakenly applied this change to all tools, responding, \textit{``The robot-assisted feeding system has been updated to use a button to complete actions when transferring drinks, utensils, and wipes.''} Despite the error, the system's transparency features allowed the user to recover. CR1 utilized transparency to verify the transfer completion method by asking, \textit{``What is the default action to complete a transfer?''} and \textit{``What other ways can I end a transfer besides pushing the button?''} After several transparency-adaptability iterations, CR1 submitted a detailed adaptability request: \textit{``Use sensors to end transactions with utensil use, the button to end transactions when taking a sip, and the button to end transactions for face wipes.''} Through these transparent interactions, CR1 was able to iteratively refine the robot's behavior to match their preferences, even when sometimes the initial adaptability command did not function as intended.

\begin{tcolorbox}[colback=itwasallyellow, colframe=darkyellow, title={\textbf{Lesson 3}}, 
fonttitle=\bfseries, 
boxsep=0.5mm, left=1mm, right=1mm, top=0.5mm, bottom=0.5mm, arc=0mm]
\linespread{0.9}\selectfont
Providing multiple interfaces is essential for transparency, as users may not always be able to interact with a single interface due to situational and environmental constraints.
\end{tcolorbox}

Over multiple days of evaluating a flexible system, users often forgot the robot’s settings and expectations (which gesture it is waiting to detect). FEAST provides transparency through voice prompts and a web interface, with multiple options proving crucial. In social settings, CR1 and CR2 switched off voice prompts and relied on the web interface for guidance. However, in cases where neither was accessible—such as when CR2 used a personalized gesture detector for \textit{``keep head still''} during mouth wiping—the wiper’s position right at their mouth prevented them from checking the interface. As CR2 noted, \textit{``A small display on the gripper would help. I can't check the web interface while the fork or another tool is at my mouth and the robot expects something. Or maybe I could ask directly, like `Hey robot, what am I supposed to do right now?'"} 

\begin{tcolorbox}[colback=itwasallyellow, colframe=darkyellow, title={\textbf{Lesson 4}}, 
fonttitle=\bfseries, 
boxsep=0.5mm, left=1mm, right=1mm, top=0.5mm, bottom=0.5mm, arc=0mm] 
\linespread{0.9}\selectfont
Cognitive workload generally decreases as users become more familiar with the system, but it also depends on the context and specific settings they choose.
\end{tcolorbox}

A flexible system allows users to tinker and personalize their experience, but it can also impose cognitive workload. While NASA-TLX surveys indicated that our system did not impose significant cognitive workload during evaluation meals, this may be because introductory meals helped users acclimate. Over three meals, cognitive workload generally trended downward (Figure~\ref{fig:study_results}) as users became more familiar with the system, except for Meal ID 6 (a social meal for CR1). CR1 noted that the mental and temporal demand was not caused by the personalization process itself but rather by the specific settings they had chosen, which were not optimal for the context. \textit{``I set the [auto-continue task selection] timer to 100 seconds, but that wasn’t enough. I kept watching the countdown while trying to keep up with the conversation, making sure I was ready for my next bite before time ran out. Next time, I’d set it much longer."} Despite this challenge, they acknowledged that familiarity would make the process easier over time: \textit{``The more I eat in this setting, the easier it’ll get. I’ll get used to what the robot can do and figure out which settings actually work for me."}

\begin{tcolorbox}[colback=itwasallyellow, colframe=darkyellow, title={\textbf{Lesson 5}}, 
fonttitle=\bfseries, 
boxsep=0.5mm, left=1mm, right=1mm, top=0.5mm, bottom=0.5mm, arc=0mm]
\linespread{0.9}\selectfont
System failures can occur in-the-wild for various reasons, but system flexibility and keeping the user in the loop improves the robot’s ability to recover.
\end{tcolorbox}

During our in-home evaluation, several unexpected skill failures occurred. For example, varying lighting conditions affected food perception, making it difficult for the robot to detect food items on the plate. However, the system provided a manual skill execution option on the web interface, allowing users to pick up undetected bites by selecting the skewering skill and a keypoint on a plate image. In cases where bite transfer failed and the robot was unaware, it would return to the plate to reacquire food, even though a food item was already on the fork. To recover, users switched to manual skill selection, skewering an empty space on the plate to retain the same food item on the fork and retry bite transfer.

Similarly, the default head nod detector—used to detect when a user was finished drinking or wiping their mouth—did not work well for CR1. When leaning forward to take a sip or wipe, they inadvertently leaned downward, which the system misinterpreted as a nod, causing the robot to move away too soon. However, because the system was flexible, CR1 was able to switch from gesture detection to a button for this task.

These examples highlight how system flexibility, combined with user involvement, allows the robot to recover from difficult to predict failures that arise in the real world.




\section{Assessing Ecological Validity with an Occupational Therapist}
\label{sec:occupational_therapist}

To assess FEAST's performance beyond the two community researchers, we evaluate our system with an Occupational Therapist (OT) not involved in system development procedure. Since OTs work with diverse care recipients with varying mobility limitations, their evaluation offers a comprehensive measure of the system’s ecological validity. The OT was introduced to FEAST as in our in-the-wild evaluations---through the training manual and a training meal.  After reviewing personalization categories from our formative study, the OT independently created 10 realistic personalization scenarios. For each scenario, they personalized the default FEAST system to their requirements using adaptability and transparency requests. We report: (1) user-reported personalization success rated at $5/5$, averaged across scenarios; (2) user-reported ease of personalization rated at $4.6 \pm 0.16$ out of 5, averaged across scenarios; (3) a preference for the personalized system over the default (no-personalization) system in all 10 scenarios; (4) a total of 13 adaptability and 2 transparency requests across all scenarios; and (5) a total of 2 experimenter interventions and 1 experimenter explanation throughout the evaluation. User ratings are on a 1–5 Likert scale (1 = Strongly Disagree, 5 = Strongly Agree). Appendix F details the personalization scenarios, including the adaptability and transparency requests made and any experimenter interventions or explanations. The OT also offered feedback on the interface design, stating: ``Visually, it looks really good; the contrast is good, it should work for most care recipients... having speech-to-text is helpful because that is how many communicate with their devices.''

\section{Limitations and Future Work}

\textbf{Open-Set Food Detection.} FEAST receives user input about food items on the plate and uses an open-set object detector to locate them. However, the detector often mislabels items, particularly when plates contain multiple food types (e.g., chicken nuggets and potato wedges). During evaluation,  additional qualifiers like ``yellow potato wedge piece'' or ``cut-up chicken nugget piece'' were required. While users could potentially provide such qualifiers to improve detection accuracy, the effectiveness of this approach remains to be evaluated. In the future, more accurate open-set detectors are needed to address this limitation.

\textbf{Adaptability Based on Plate Context}: Some user adaptability requests required the system to have additional context about plate contents and selectively adjust behavior tree parameters. For instance, in Meal ID 1, CR1 requested more ranch for potato wedges and less for chicken bites. However, since FEAST currently uses a single dipping parameter, this required constant adjustments between bites. Future work should explore adaptability mechanisms that dynamically adjust behavior based on real-time plate contents.

\textbf{LLM Hallucinations.} FEAST leverages LLMs to interpret user requests and modify robot behavior, but hallucinations can lead to incorrect or incomplete adaptations, causing user confusion. While FEAST’s transparency features help users identify and correct such errors, these features themselves rely on LLMs, which may also hallucinate. Future research should focus on reducing LLM-induced errors and exploring fallback mechanisms to ensure reliability.

\textbf{Long-term Preference Learning}: CR1 noted that generating a full set of personalization requests from scratch for a new setting could impose unnecessary cognitive workload. They suggested allowing users to describe their dining environment so that the system could initialize a relevant configuration, from which they could make refinements. Achieving this would require long-horizon preference learning, where FEAST analyzes past user preferences to propose suitable initial settings for different contexts.

\textbf{Evaluation Beyond Community Researchers.} Our evaluation involved two care recipients who are also community researchers. Since they were familiar with FEAST’s development, they may have had a lower cognitive workload than other users. Future studies should evaluate FEAST with care recipients who are not community researchers to better understand how unfamiliar users perceive the system’s usability.

\textbf{Open-Loop Skill Limitations.} FEAST employs open-loop skills for bite acquisition, which can fail when food items slip. Moreover, while users can adjust skill parameters to control bite size—such as scooping more or less—these settings often require frequent adjustments within a single meal as food properties change over time. For example, the same scooping distance may yield smaller portions as food consistency shifts. Incorporating closed-loop approaches to bite acquisition could improve robustness, allowing the system to adapt dynamically to food slippage and changing food properties.








\section{Acknowledgement}

All data collection and user studies described in this paper were approved by the Cornell University Institutional Review Board. This work was partly funded by the National Science Foundation under grants IIS \#2132846 and CAREER \#2238792. We are deeply grateful to Aimee Whitneck (CR1) and Benjamin Howe (CR2) for their involvement throughout this work. We also thank Amal Nanavati for assisting with participant recruitment for the formative user study, and all formative study participants for their valuable feedback, which guided system development. Finally, we thank the anonymous reviewers for their insightful feedback.

\bibliographystyle{IEEEtran}
\bibliography{references/feeding,references/deployment,references/personalization,references/large_language_models,references/system}

\newpage

\twocolumn[
\centering
\Huge{Appendix} \\
\Large{\textbf{FEAST}: A \underline{\textbf{F}}lexible M\underline{\textbf{e}}altime-\underline{\textbf{A}}ssistance  \underline{\textbf{S}}ystem \underline{\textbf{T}}owards In-the-Wild Personalization} \\
\href{http://emprise.cs.cornell.edu/feast}{http://emprise.cs.cornell.edu/feast} 
\vspace{1em} 
]

\setcounter{section}{0} 
\renewcommand\thesection{\Alph{section}}
\renewcommand\thesubsection{\Alph{section}.\arabic{subsection}} 

\section{Caregiving Requirements Identified from Formative Study}

Refer to Table \ref{table:appendix_personalization_needs}. 

\setlength{\LTcapwidth}{\textwidth} 


\definecolor{mealtime_tasks}{HTML}{bad0d5}
\definecolor{interaction}{HTML}{e0f2eb} 
\definecolor{hardware}{HTML}{c0dee0}
\definecolor{acquisition}{HTML}{f8f3e5}
\definecolor{transfer}{HTML}{f0d2bb}
\definecolor{misc}{HTML}{fae1bf}


\tiny
\begin{table*}[t]
\centering
\caption{Key personalization requirements for in-the-wild mealtime assistance (Part I: Requirements R1--R23). Identified from a formative study with 19 care recipients and 2 community researchers, along with a comparison of how FEAST and prior feeding systems (Nanavati et al.~\cite{nanavati2023design} and FLAIR~\cite{jenamani2024flair}) address them.}
\label{table:appendix_personalization_needs}
\begin{tabular}{p{0.3cm} p{2.8cm} p{4cm} 
>{\centering\arraybackslash}p{0.8cm} 
>{\centering\arraybackslash}p{0.8cm} 
>{\centering\arraybackslash}p{0.8cm} 
p{5.7cm}}
\hline
\multirow{2}{*}{\textbf{\textsf{ID}}} &
\multirow{2}{*}{\textbf{\textsf{Requirement}}} & 
\multirow{2}{*}{\textbf{\textsf{Description}}} & 
\multirow{2}{*}{\textbf{\textsf{FEAST}}} & 
\textbf{\textsf{Nanavati et al.}} & 
\multirow{2}{*}{\textbf{\textsf{FLAIR}}} & 
\multirow{2}{*}{\textbf{\textsf{Explanation}}} \\
\hline
\rowcolor{mealtime_tasks} 
& \textbf{\textsc{Mealtime Tasks}} & & & & & \\
\rowcolor{mealtime_tasks}
R1 & Feeding 
    & Assist with eating solid food.
    & \cmark & \cmark & \cmark
    & All three systems can perform feeding tasks. \\
\rowcolor{mealtime_tasks}
R2 & Drinking 
    & Assist with consuming liquids.
    & \cmark & \xmark & \xmark
    & Only FEAST supports drinking. \\
\rowcolor{mealtime_tasks}
R3 & Static Mouth Wiping 
    & Assist with moving a wiper to near the user's mouth.
    & \cmark & \xmark & \xmark
    & Only FEAST supports static mouth wiping. \\
\rowcolor{mealtime_tasks}
R4 & Dynamic Mouth Wiping 
    & Dynamically wipe a user's mouth who remains stationary.
    & \xmark & \xmark & \xmark
    & None of the systems support this. \\
\rowcolor{mealtime_tasks}
R5 & Manual Task Selection 
    & Allows the user to choose which mealtime task (feeding, drinking, wiping) to perform.
    & \cmark & - & -
    & Nanavati et al.\ and FLAIR only consider feeding. \\
\rowcolor{mealtime_tasks}
R6 & Auto-Execute Last Task
    & Repeat the previous mealtime action if re-selection does not occur until a certain timeout.
    & \cmark & - & -
    & Nanavati et al.\ and FLAIR only consider feeding. \\
\rowcolor{mealtime_tasks}
R7 & Auto-Detect Mouth Wiping Requirement 
    & Perceive residue on users mouth to trigger mouth wiping. 
    & \xmark & \xmark & \xmark
    & None of the systems support this. \\
\rowcolor{mealtime_tasks}
R8 & Meal Heating 
    & Keeps food warm or at a desirable temperature.
    & \xmark & \xmark & \xmark
    & None of the systems support this. \\
\hline
\rowcolor{interaction}
& \multicolumn{2}{l}{\textbf{\textsc{Interaction: Human to Robot}}} & & & & \\
\rowcolor{interaction}
R9  & Web Interface 
    & Provides a screen-based control option accessible via a web browser.
    & \cmark & \cmark & \xmark
    & Both FEAST and Nanavati et al.\ can be operated via a web interface. FLAIR does not one. \\
\rowcolor{interaction}
R10 & Gesture-Based Control 
    & Supports head gestures for interactions.
    & \cmark & \cmark & \cmark
    & FLAIR and Nanavati et al.\ use an open-mouth gesture to signal readiness for a bite. FEAST uses open-mouth or head-nod gestures by default. \\
\rowcolor{interaction}
R11 & Personalized Gestures 
    & Enables customization of gestures based on individual user capabilities and preferences, for example, mouth open with a smaller threshold, head shake, blinking, etc.
    & \cmark & \xmark & \xmark
    & FEAST allows adding custom gestures by providing a language description, a few positive examples, and a few negative examples, leveraging LLM-based code synthesis. \\
\rowcolor{interaction}
R12 & Button-Based Control 
    & Supports a button for interactions.
    & \cmark & \xmark & \xmark  
    & Only FEAST supports a non e-stop button.\\
\rowcolor{interaction}
R13 & Personalized Button Clicks 
    & Users request different button click patterns (e.g., three quick presses, one long press) to correspond to different aspects of the system.
    & \xmark & \xmark & \xmark
    & None of the systems support this, but the code synthesis capabilities of FEAST given positive and negative examples could be extended to support this. \\
\rowcolor{interaction}
R14 & Eye Gaze 
    & Supports eye gaze for interactions.
    & \xmark & \xmark & \xmark  
    & None of the systems support this.\\
\rowcolor{interaction}
R15 & Voice 
    & Supports voice for interactions.
    & \cmark & \cmark & \xmark  
    & Both FEAST and Nanavati et al. feature web interfaces which support speech-to-text / speech-based control.\\
\hline
\rowcolor{interaction}
& \multicolumn{2}{l}{\textbf{\textsc{Interaction: Robot to Human}}} & & & & \\
\rowcolor{interaction}
R16 & Web Interface 
    & Allows the system to present information visually on a browser.
    & \cmark & \cmark & \xmark
    & FEAST and Nanavati et al.\ display information on a web interface; FLAIR does not. \\
\rowcolor{interaction}
R17 & Voice Feedback 
    & Speaks out the robot's status or actions.
    & \cmark & \cmark & \cmark
    & All systems can provide fixed spoken prompts. Moreover, FEAST can generate feedback adaptable to current system behavior. \\
\rowcolor{interaction}
R18 & Audible Beeps 
    & Offers simple sound-based signals if voice feedback is undesirable.
    & \cmark & \xmark & \xmark
    & FEAST can use beeps to signal robot readiness; the other two systems do not. \\
\rowcolor{interaction}
R19 & LED Indicator 
    & Provides a visual cue of robot state or readiness.
    & \cmark & \xmark & \xmark
    & FEAST features an LED near the camera mount that can be customized to indicate readiness. \\
\hline
\rowcolor{hardware}
& \textbf{\textsc{Hardware}} & & & & & \\
\rowcolor{hardware}
R20 & User-Preferable Tools 
    & Enables users to choose eating tools that suit their preferences or physical needs, such as smaller tips for those with narrow mouth openings.
    & \cmark & \xmark & \cmark 
    & FEAST supports switchable-tip silicone and metal utensils, a Velcro-attached replaceable mouth wiper, and an adaptable handle for user-provided mugs with fixed straws. Nanavati et al.\ use a fixed 3D-printed fork, while FLAIR also supports switchable-tip silicone or metal utensils. \\
\rowcolor{hardware}
R21 & Autonomous Tool Change 
    & Reduces the need for caregiver intervention when switching between tools for different tasks.
    & \cmark & \xmark & \xmark 
    & FEAST can switch between tools (feeding utensil and mouth wiper) in its mount and also pick up a drink from a table. \\
\rowcolor{hardware}
R22 & Detachable Tool Tips 
    & Facilitates easier cleaning, maintenance, or replacement of utensils.
    & \cmark & \xmark & \cmark 
    & FEAST and FLAIR tool tips can be removed easily for cleaning. Nanavati et al. use a 3D-printed metal utensil that must be intricately screwed onto the mount with three screws. \\
\rowcolor{hardware}
R23 & Flexible Mounting 
    & Allows system installation on various supports (e.g., a wheelchair or stand) for user convenience.
    & \cmark & \cmark & \cmark 
    & FEAST mounts on a wheelchair or Vention stand, Nanavati et al.\ on a wheelchair or movable table, and FLAIR on a Vention stand or table. \\
\hline
\end{tabular}
\end{table*}
\normalsize


\tiny
\begin{table*}[t]
\centering
\caption{Key personalization requirements for in-the-wild mealtime assistance (Part II: Requirements R24--R46).}
\label{table:appendix_personalization_needs_part2}
\begin{tabular}{p{0.3cm} p{2.8cm} p{4cm} 
>{\centering\arraybackslash}p{0.8cm} 
>{\centering\arraybackslash}p{0.8cm} 
>{\centering\arraybackslash}p{0.8cm} 
p{5.7cm}}
\hline
\multirow{2}{*}{\textbf{\textsf{ID}}} &
\multirow{2}{*}{\textbf{\textsf{Requirement}}} & 
\multirow{2}{*}{\textbf{\textsf{Description}}} & 
\multirow{2}{*}{\textbf{\textsf{FEAST}}} & 
\textbf{\textsf{Nanavati et al.}} & 
\multirow{2}{*}{\textbf{\textsf{FLAIR}}} & 
\multirow{2}{*}{\textbf{\textsf{Explanation}}} \\
\hline
\rowcolor{acquisition}
& \textbf{\textsc{Acquisition}} & & & & & \\
\rowcolor{acquisition}
R24 & Bite Prediction 
    & Automated selection of which food item to acquire next based on user preferences.
    & \cmark & \xmark & \cmark
    & FEAST and FLAIR use a bite-ordering preference (e.g., “feed all of X, then Y”) for bite prediction. Nanavati et al.\ require manual bite selection. \\
\rowcolor{acquisition}
R25 & Auto-Continue Predicted Bite 
    & Automatically continue with predicted next bite after a timeout.
    & \cmark & \xmark & \cmark
    & Nanavati et al.\ do not predict bites, so they cannot auto-continue them. \\
\rowcolor{acquisition}
R26 & Manual Bite Selection 
    & Allows the user to specifically choose which piece of food to pick up.
    & \cmark & \cmark & \xmark
    & FLAIR does not feature any user interface for bite selection. Users can override FEAST's autonomously predicted bite via the web interface. \\
\rowcolor{acquisition}
R27 & Diverse Fork-Based Acquisition Skills 
    & Supports various acquisition (skewering, scooping, twirling, dipping, etc.) and pre-acquisition actions (pushing, grouping, cutting, etc.).
    & \cmark & \xmark & \cmark
    & Both FLAIR and FEAST can skewer, scoop, dip, and perform limited cutting of soft foods. Nanavati et al.\ only support skewering. \\
\rowcolor{acquisition}
R28 & Finger Food Acquisition Skills 
    & Supports picking up finger foods.
    & \xmark & \xmark & \xmark
    & None of the systems support this. \\
\rowcolor{acquisition}
R29 & Bite Size Customization
    & Enables adjustments to the portion size for each bite.
    & \cmark & \xmark & \xmark
    & FEAST allows commands like “Scoop more” or “Dip more” to change bite size for scooping or dipping. For skewering, it autonomously selects smaller food pieces. Caregivers must provide bite-sized portions for hard items. \\
\rowcolor{acquisition}
R30 & Skewering Axis Selection 
    & Allows adjusting fork orientation to tradeoff between easy food acquisition and transfer through narrow mouth openings.
    & \cmark & \xmark & \xmark
    & FEAST enables users to change the skewering axis via adaptability commands. \\
\rowcolor{acquisition}
R31 & Ask Confirmation After Acquisition 
    & Allows users to confirm moving to transfer after acquisition.
    & \cmark & \cmark & \xmark
    & FEAST lets users enable or disable confirmation prompts. Nanavati et al.\ always ask for confirmation with a fixed auto-continue time. \\
\hline
\rowcolor{transfer}
& \textbf{\textsc{Transfer}} & & & & & \\
\rowcolor{transfer}
R32 & Adjust Outside-Mouth Distance 
    & Lets users define how close the tool should come to their mouth for outside-mouth transfers.
    & \cmark & \cmark & \xmark
    & Users can specify this distance in FEAST via adaptability commands. Nanavati et al.\ adjust it in a settings page. \\
\rowcolor{transfer}
R33 & Adjust Inside-Mouth Placement 
    & Allows customizing how far the utensil enters the mouth for comfort.
    & \cmark & \xmark & \cmark
    & Both FEAST and FLAIR let users calibrate the inside-mouth position. Nanavati et al.\ do not do inside-mouth transfer. \\
\rowcolor{transfer}
R34 & Convey Robot Ready to Initiate Transfer 
    & Notify user when robot is ready to intiatate transfer
    & \cmark & \cmark & \cmark  
    & All systems support this, but FEAST offers more interchangeable interfaces (e.g., voice, LED). \\
\rowcolor{transfer}
R35 & Detect User Ready for Transfer 
    & Perceive if the user is ready to initiate transfer
    & \cmark & \cmark & \cmark  
    & All systems support this, but FEAST offers more interchangeable interfaces (e.g., button, gestures).\\
\rowcolor{transfer}
R36 & Convey User can Take Bite / Sip / Mouth Wipe
    & Notify user that they can take complete transfer
    & \cmark & \cmark & \cmark  
    & All systems support this, but FEAST offers more interchangeable interfaces (e.g., voice, LED). \\
\rowcolor{transfer}
R37 & Detect Transfer Complete
    & Perceive that the user has completed transfer
    & \cmark & \xmark & \cmark  
    & All systems support this, but FEAST offers more interchangeable interfaces (e.g., force-torque sensing for detecting bites, button, gestures). \\
\hline
\rowcolor{misc}
& \textbf{\textsc{Miscellaneous}} & & & & & \\
\rowcolor{misc}
R38 & Speed Control 
    & Allows adjusting the robot's movement speed for safety or efficiency.
    & \cmark & \cmark & \xmark
    & FEAST uses language commands to update speed globally or for specific skills. Nanavati et al.\ can change bite transfer speed in settings. \\
\rowcolor{misc}
R39 & Non-Obtrusive Movement 
    & Minimizes intrusion into the user's personal space or field of view.
    & \cmark & \xmark & \xmark
    & FEAST’s custom tools can be held from the side, like humans do, reducing occlusion compared to Nanavati et al.\ and FLAIR. \\
\rowcolor{misc}
R40 & Socially Acceptable Movement 
    & Ensures the robot’s motions appear natural and comfortable in social settings.
    & \xmark & \xmark & \xmark
    & None of the systems explicitly address social acceptability of movement. \\
\rowcolor{misc}
R41 & Retract From Certain Configurations 
    & Allows the user to retract the robot from certain poses for safety or convenience.
    & \cmark & \xmark & \xmark
    & FEAST lets users add a retract node via adaptability commands at specific arm configurations. \\
\rowcolor{misc}
R42 & Robot Voice Prompts 
    & Enables custom voice prompts for guidance or reminders.
    & \cmark & \xmark & \xmark
    & FEAST allows adding custom voice prompts at any node through adaptability commands. \\
\rowcolor{misc}
R43 & Long-Term Preference Learning 
    & Learn user preferences over time to improve assistance.
    & \xmark & \xmark & \xmark
    & None of the systems support this. \\
\rowcolor{misc}
R44 & Adjust Fixed Configurations 
    & Allows editing fixed configurations (e.g. above plate config).
    & \xmark & \cmark & \xmark
    & Nanavati et al.\ let users adjust fixed configurations; Others require experimenters to do so. \\
\rowcolor{misc}
R45 & Adapt to User-Robot-Plate Setup 
    & Ensures the system can function with different user seating and plate placements.
    & \cmark & \cmark & \cmark
    & All systems adapt to variations in user-robot-plate arrangement leveraging perception. \\
\rowcolor{misc}
R46 & Scene Selection
    & Enables recalling and switching between saved settings for different scenes.
    & \xmark & \cmark & \xmark
    & Nanavati et al.\ let users switch between pre-configured scenes. FLAIR does not consider different scenes. Users can adapt FEAST to new scenes, but need experimenter help to load settings from prior scenes. \\
\hline
\end{tabular}
\end{table*}
\normalsize

\section{Gesture Synthesis using Large Language Models}

We now describe our custom gesture synthesizer in more detail. The synthesizer takes a natural language description of the gesture (e.g., “shaking my head from left to right”), 3-5 positive examples, and 3-5 negative examples supplied by the user. Each of these training examples comprises a time series of head poses and facial key points, which are obtained from the our head perception. The user controls data collection through the interface. In practice, the examples are sampled at a rate of roughly 10Hz and typically last for 5-10 seconds.

Given these inputs, we use an LLM (GPT-4o~\cite{hurst2024gpt}) to generate a Python function capable of detecting specific gestures. The Python function takes input from the head perception model, external events that may shut down the detector, and a timeout counter; it outputs if the gesture has been detected. We prompt the LLM with two in-context examples: one for detecting “mouth open”, and the other for detecting “head nod”. We also provide a description of the head perception module that is available to the Python function. Each Python function proposed by the LLM is validated with respect to the positive and negative examples. If evaluating any of the examples raises an exception, or if the Python classification misclassifies the example, we automatically re-prompt the LLM up to 3 times with failure information before returning the best-seen program.

As an additional extension, we allow the LLM to propose real-valued hyperparameters for the Python function detector. For example, in the case of “shaking my head from left to right”, the LLM automatically proposes a hyperparameter for head yaw angle from left to right, \textit{$head\_shake\_threshold$}. Following the in-context examples, each hyperparameter is proposed along with a lower and upper bound. We then optimize these hyperparameters for each proposed program using a grid search with 10 values per dimension. The objective for optimization is classification accuracy with respect to the user-provided positive and negative examples.

Once synthesized, the gesture can be tested by the user through the interface and added to behavior trees through the natural-language-based personalization procedure described in the main paper. When the generated function is invoked, it is dynamically executed within a separate memory-controlled-workspace for safety.

To evaluate FEAST's gesture personalization capabilities against a baseline that doesn't use personalized parameters, we conduct an ablation study. We compile 8 gesture types from the formative study:
\begin{itemize} \item Head nod (up-down) \item Head shake (left-right) \item Head tilt (left-right) \item Head still (5 seconds) \item Mouth open \item Mouth continuously open (5 seconds) \item Three successive eye blinks \item Three successive eyebrow raises \end{itemize}
Three individuals without mobility limitations emulate each gesture over a prescribed range of motion while secured to a wheelchair with restraint bands---a technique OTs routinely use to simulate mobility limitations. These emulations are post-validated by an occupational therapist referencing Zoom-recorded snippets from our formative study. Following the FEAST setup, we collect 5 positive and 5 negative examples per individual and gesture, along with corresponding language descriptions, to synthesize gesture detectors with personalized parameters. The personalized gesture detector achieves an F1-score of $0.9 \pm 0.02$ on the examples, compared to $0.63 \pm 0.09$ for no personalized parameters.
\section{Adherence to ISO 13842:2014}

Transferring food, drinks inside the mouth and wiping requires nuanced physical interactions. These physical interactions, if completed in an unsafe manner, may lead to hazards during human-robot collaboration. The ISO 13842:2014, Robots and robotic devices — Safety requirements for personal care robots, is a guideline developed by the International Organization for Standardization to identify possible hazards and address them. We show that our mealtime-assistance system adheres to 13842:2014.

Our system uses a Kinova Gen3 7DoF which, per its user guide, adheres to ISO / TS 15066 Collaborative robots certification, ISO 12100 Safety of machinery, ISO 13849 Safety-related parts of control systems, ISO 10218-1 Industrial robots, IEC 60204-1 Safety of machinery – Electrical, and IEC 60950-1: Information Technology Equipment – Safety. Some of these are also requirements in 13842:2014.

Below is a summary of all the robot's safety features. We will reference them when explaining how we satisfy the requirements listed in 13842:2014. We will refer to \textit{pose filtering} as \textbf{A}, \textit{compliant control} as \textbf{B}, \textit{active collision monitoring} as \textbf{C}, \textit{safety-centric hardware design} as \textbf{D}, \textit{emergency stops} as \textbf{E}, and \textit{watchdog} as \textbf{F}. These are described in detail in the paper.

\subsection{ISO Guidelines} 
\textbf{4 Risk assessment:} \\
\textbf{4.1 General:} The system adheres to ISO 12100 as per the Kinova Gen3 7DoF user guide.
\textbf{4.2 Hazard identification \& 4.3 Risk estimation:}
\textbf{(a)} uncertainty of autonomous decisions made by the robot and possible hazards from wrong decisions; Autonomous decisions made by the robot are the movements to pickup food, acquisition, and path the robot takes to feed an operator. \textbf{A} reduces hazards a controlled zone. \textbf{C} monitors for collisions with this zone, further reducing risk of danger to operator. \textbf{(b)} different levels of knowledge, experience and physical conditions of users and other exposed persons; Will provide transparency so that users know the details about how the robot works as well as all safety features at the beginning of each study. May have varying physical conditions that our robot can adjust too. \textbf{(c)} normal but unexpected movement of the personal care robot; We will demonstrate robot's movements to user pre-experiment to minimize this. \textbf{(d)} unexpected movement (e.g. jumping in front of the personal care robot from the side or from a higher level) of humans, domestic animals and other safety-related objects; During velocity control, we have \textbf{C} to stop the robot in the case of unexpected movements. During \textbf{B}, users can push the robot with minimal force to move it away. \textbf{(e)} unintended movement of the personal care robot; Unintended movements of the robot would be in the pre-defined zone by \textbf{A} and robot system has \textbf{C} and \textbf{B}. \textbf{(f)} unexpected travel surfaces and environmental conditions in the case of mobile robots; The system has two options for mounting. It can be mounted to a wheelchair system or to a vention stand. For vention stands, operator dictates movement by hand. For wheelchair, as products in the market, they must satisfy the ISO 7176 which regulates wheelchair safety. \textbf{(g)} uncertainty of safety-related objects to be handled in the case of mobile servant robots; \textbf{C} \& \textbf{B} \textbf{(h)} conformity to the human anatomy and its variability in the case of physical assistant robots and person carrier robots; System detects head poses and dimensions, creating a custom model that fits the user that is currently being fed.

\textbf{5 Safety requirements and protective measures:} \\
\textbf{5.1 General:} \textbf{(a)} inherently safe design; \textbf{(b)} protective measures; \textbf{(c)} information for use; are explained in \textbf{(5.2.2)}, \textbf{(5.2.3)}, \textbf{(5.2.4)}. \textbf{5.2 Hazards related to charging battery:} \textbf{(5.2.1)} The Kinova Gen3 7DoF robotic arm operates on an external power supply not a built-in battery charging system. Thus we do not need to satisfy requirements mentioned in \textbf{5.2.1}. \textbf{(5.2.2)} As per the Kinova Gen3 7DoF user guide, the system satisfies IEC 60204-1. All live parts are covered in the system. \textbf{(5.2.3)} \textbf{(a)} charging systems shall be designed in such a way that the charging connections are only activated when the personal care robot is connected to them; Battery and charging are all external and not in the robotic arm. \textbf{(b)} batteries show status of charge, and that charging connections are only activated if the personal care robot is connected. \textbf{(5.2.4)} Provided in the Kinova 3Gen 7DoF user guide and in the user guide for the operator's wheelchair if arm is connected to one. \textbf{5.3 Hazards due to energy storage and supply:} All parts of \textbf{5.3} are satisfied as (a) the system uses a external battery, (b) system does not have pneumatic equipment or hydraulic equipment, (c) the system follows IEC 60204-1, IEC 60950-1, ISO 10218-1, and ISO 12100. The aforementioned IECs and ISOs are extensions or incorporate all the IEC and ISO requirements mentioned in \textbf{5.3}. \textbf{5.4 Robot start-up and restart of regular operation:} \textbf{(5.4.1)} System has two start-ups. One is hardware start-up and the other is the feeding algorithm start-up. On both start-ups, the robot stays in a neutral position away from any operator. \textbf{(5.4.2)} \textbf{(a)} During start-up the personal care robot shall perform internal control tests to ensure that all safety-related functions are available. Failure to do so shall avoid any hazardous operation: Kinova arm does internal control tests and won't move if anything fails. Our feeding system also has the \textbf{F} to check other safety features and on start-up a script also checks to ensure \textbf{F} is running. \textbf{(b)} If the personal care robot’s safety-related functions cannot be performed correctly after start up, then it shall immediately perform a protective stop: If the script detects that \textbf{F} is not running, robot doesn't start up. If \textbf{F} finds any other safety features are not working, stops the robot and enters a safe state. \textbf{(c)} The personal care robot shall start up in a state of restricted speed, force, etc. (see 6.4 and 6.7) and shall only return to normal levels of control by means of a mode change (see 6.11).: System follows 6.4, 6.7, and 6.11. \textbf{(d)} The personal care robot shall always start up in manual mode, and shall only continue operation in autonomous mode by means of a mode change as specified in 6.11.1: System follows 6.11.1. \textbf{(5.4.3)} \textbf{(a)} Manipulators, mobile platforms and other moving parts shall be de-activated by safety-related
functions at start-up (to prevent any unintended actions). Application functions shall only be enabled if it can be positively established, via sensors, that no hazardous situation exists. This measure shall be applied if the robot is intended to enter autonomous mode immediately after start-up: Our system does (a). \textbf{(b)} The personal care robot shall always start up in a monitored standstill and shall only return to normal operation by means of a user action: System also satisfies (b). \textbf{5.6 Hazards due to robot shape:} \textbf{(5.6.1 \& 5.6.2)} Robot system satisfies ISO 12100, ISO 13854, and ISO 15534 as per the user guide. \textbf{(5.6.3)} \textbf{(a)} cushioning on sharp edges and points shall be provided to eliminate shearing, stabbing, cutting
hazards and reduce impact hazards (see 4.3, Note 2): \textbf{D} satisfies this. \textbf{(b)} use of fixed or movable guards to cover hazardous moving parts: Done as per Kinova user guide. Added a dynamixel motors, but these are compliant and can be easily forced around. \textbf{(c)} adjustment of the robot speed and behaviour, if hazardous loads (e.g. sharp or pointed objects) are being carried: \textbf{F} monitors speed, force, etc., keeping them at safe levels, at all times. \textbf{(5.6.4)} system adheres to ISO 12100 and ISO 7010. \textbf{5.7 Hazards due to emission \& 5.8 Hazards due to electromagnetic interference:} Kinova satisfies requirements in \textbf{5.7} \& \textbf{5.8} and our added features due not release new emissions or add new features that would be impacted due to electromagnetic interference. \textbf{5.9 Hazards due to stress, posture and usage:} \textbf{5.9.2} addresses physical while \textbf{5.9.3} addresses mental stress. \textbf{(5.9.2.1)} \textbf{(a)} A personal care robot shall be designed to minimize or reduce physical stress or strain to its user due to continuous use, including but not limited to uncomfortable posture, operational environments that would directly cause physical discomfort, e.g. fatigue and tendon inflammation: We allow users to specify inside mouth or outside mouth bite transfer, with the algorithm being tuned to their preferences, to maximize comfort. \textbf{(b)} The design of the personal care robot shall take into account typical body sizes of the intended user population in order to avoid physically demanding body postures or to ensure easy operation. ISO 14738 describes how principles of ergonomic factors, by construction of workstations and machinery, should be applied. This should be considered when building a personal care robot, where somebody sits on or stands in front of the robot: System design ISO 14738 into account and has \textbf{A} to account for human variability in comfort. \textbf{(5.9.2.2)} \textbf{(a)} the design and location of manual control devices, which should ensure that they can be operated without physical stress or discomfort; Emergency stop buttons are close to user and take minimal force to press. \textbf{(b)} the proper ergonomic design and location of the seat, which should ensure that good posture can be maintained during operation of the personal care robot: We attach to user's wheelchair, so it should already be adjusted to user's comfort. \textbf{(c)} command devices that are detachable or hand-held instead of being permanently attached to the personal care robot in an inappropriate position: \textbf{E} is detachable and one is velcroed within immediate reach of the user while the other is always close to the experimenter so that both can reach in an emergency. User's web interface's smart device is attached to a stand that will connect to the wheelchair. \textbf{(5.9.3.1)} \textbf{(a)} a personal care robot shall be designed to minimize or reduce mental stress to its user due to continuous use;
Robot is controled via a user web-interface where buttons are clicked to control the robot's next move. The user's web-interface was designed so that it's interactive buttons are large and easily to click. \textbf{(b)} user interfaces such as controls, signalling or data display elements, shall be designed to be easily understood so that clear and unambiguous interaction between the human and the personal care robot is possible: Our user interface's buttons is clearly labeled in large font. \textbf{(c)} the personal care robot shall conform to ergonomics standards appropriate to its intended purpose (see ISO/TR 9241-100, ISO 9241-210, ISO 9241-400, ISO 9241-920 and ISO 11228): Conforms to the above ISOs. \textbf{(5.9.3.2)} Design of the user interface was created with user's in loop, so user's know what to expect in terms of signals, display design, etc. \textbf{5.10 Hazards due to robot motion:} \textbf{(5.10.2.1, 5.10.2.2, \& 5.10.2.3)} Kinova is mechanically stable and follows the inherently safe design points in \textbf{5.10.2.2}. Have \textbf{F} which satisfies the protective measures mentioned in \textbf{5.10.2.3}. \textbf{(5.10.3)} The user's wheelchair should adhere to the requirements about instability during travel discussed in \textbf{5.10.3}. \textbf{(5.10.4)} There is no instability while carrying the fork and food in velocity control mode. In \textbf{B}, we appropriately change the dynamic properties of the robot arm, so that there is no hazards while carrying the utensils and food. \textbf{(5.10.5.1)} \textbf{(a)}  Risk assessment shall determine the allowable maximum values of the appropriate parameters (e.g. contact force) that have an influence on the risk caused by contact over the entire operational range: covered in ISO 12100 and ISO 10218-1. \textbf{(b)} A personal care robot shall be designed to ensure that it does not cause any hazardous rollovers,runaways, or detachment of robot body parts, even when it receives any collision forces or safety-related obstacle detection signals during its motion, up to the limits specified for its intended operation: covered in ISO 12100 and ISO 10218-1. \textbf{(5.10.5.2 \& 5.10.5.3)} Implementations to Kinova arm that satisfy requirements in \textbf{5.10.5.2 \& 5.10.5.3} can be found in Kinova 3Gen 7DoF user guide. Furthermore, we have \textbf{C} that stops robot when anomally is detected. \textbf{(5.10.6)} Irrelevant as robot is not a restraint type physical assistant robot. \textbf{(5.10.7)} Irrelevant as robot is not a person carrier robot. \textbf{(5.10.8 \& 5.10.9)} The system has \textbf{F} that monitor and control velocity/force and \textbf{C} that watches for obstacles in surrounding area to minimize collisions. All requirements listed in \textbf{5.10.8 \& 5.10.9} are satisfied by these safety features. Furthermore, the only safety-related obstacle is the user in our system environment and robot is only near user when doing bite, drink transfer or mouth wiping which is when the robot is in \textbf{B}. \textbf{(5.10.9)} Information to satisfy requirements in \textbf{5.10.9} can be found in Kinova user guide. \textbf{5.11 Hazards due to insufficient durability:} Information to satisfy requirements in \textbf{5.11} can be found in Kinova user guide. \textbf{5.12 Hazards due to incorrect autonomous decisions and actions:} \textbf{(5.12.1)} System can correctly detect what is food and what is not food. Has powerful force sensors, velocity sensors, etc. that are closely monitored to keep hazards during autonomous movements to a minimum. \textbf{(5.12.2)} \textbf{(a)} constraining the operational scenarios to reduce risk of harm due to incorrect actions: \textbf{A} satisfies (a). \textbf{(b)} use of unique identifiers for safety-related objects, travel paths, etc: \textbf{A} also satisfies (b). \textbf{(5.12.3)} The system uses multiple checks to make sure a autonomous decision is correct and allows user input to stop any process if something goes around. \textbf{5.13 Hazards due to moving components:} Requirements here are satisfied by \textbf{A}, \textbf{C}, and \textbf{E}. Furthermore, \textbf{D} makes sure that even in the instance of a collision, little harm is done to a person. \textbf{5.14 Hazards due to lack of awareness of robots by humans:} Points listed in \textbf{5.13} and detector for when a user may not be paying attention to the robot when some form of transfer is occurring satisfying requirements to \textbf{5.14}. \textbf{5.15 Hazardous environmental conditions \& 15.6 Hazards due to localization and navigation errors:} Satisfied by the points listed in \textbf{5.13}.

\textbf{6 Safety-related control system requirements:}
\textbf{6.1 General:} Requires \textbf{(a)} emergency stop: Kinova arm provides a emergency stop (experimenter only), \textbf{(b)} protective stops: \textbf{F} \& \textbf{E} (both user and experimenter), \textbf{(c)} limits to workspace: \textbf{A}, \textbf{(d)} safety-related speed control: \textbf{F}, \textbf{(e)} safety-related force control: \textbf{F}, \textbf{(f)} hazardous collision avoidance: \textbf{C} \& \textbf{B}, and \textbf{(g)} stability control (including overload protection): description for these can be found in Kinova user guide \& \textbf{D}. \textbf{6.1.2} Irrelevant as robot is not restraint type or person carrier type. \textbf{6.1.3} System is a physical assistant robot of type 2.4, with safety features listed above satisfying the type requirements \textbf{(d)} in \textbf{6.2.2.2}, \textbf{(c)} in \textbf{6.2.2.3}, \textbf{(d)} in \textbf{6.3}, \textbf{(d)} in \textbf{6.4}, \textbf{(b)} in \textbf{6.7}, and \textbf{(d)} in \textbf{6.5.2.1, 6.5.2.2, 6.6, \& 6.7}.

\section{In-the-wild Evaluation Questions}

\subsection{Pre-Study Questionnaire.} 
\begin{itemize}
  \item What is your age?
  \item What is your gender?
  \item What is your race?
  \item What is your medical condition?
  \item How long have you had this medical condition?
  \item What are your daily challenges with getting fed by your caregiver?
  \item Do you have any familiarity with robotic/assistive technologies? If yes, please share your experience.
\end{itemize}

\subsection{Pre/post study caregiving comparison (1 - 5 scale, very low to very high).} 
\begin{itemize}
  \item \textbf{Effort:} I have to put in significant effort to communicate my preferences during meals when assisted by my caregiver/the robot.
  \item \textbf{Control:} I feel in control of my feeding experience when assisted by my caregiver/the robot.
  \item \textbf{Independence:} I feel a sense of independence when I receive assistance from my caregiver/the robot.
\end{itemize}

\subsection{NASA TLX (1 - 7 scale, very low to very high).} 
\begin{itemize}
  \item \textbf{Mental Demand:} How mentally demanding was using the meal-assistance system?
  \item \textbf{Physical Demand:} How physically demanding was using the meal-assistance system?
  \item \textbf{Temporal Demand:} How much time pressure did you feel when using the meal-assistance system?
  \item \textbf{Performance:} How successful were you in using the meal-assistance system to achieve your goals? (We flip this while reporting scores.)
  \item \textbf{Effort:} How hard did you have to work to use the meal-assistance system effectively?
  \item \textbf{Frustration:} How frustrated were you while using the meal-assistance system?
\end{itemize}

\subsection{Technology Acceptance Model (1 - 5 scale, Strongly Disagree - Strongly Agree).} 
\begin{itemize}
  \item \textbf{Perceived Usefulness:} Using this meal-assistance system will make me more independent in eating.
  \item \textbf{Perceived Ease of Use:} This meal-assistance system is easy to use.
  \item \textbf{Attitude Toward Using:} Using the meal-assistance system for improving my independence is a good idea.
  \item \textbf{Behavioral Intention to Use:} Assuming I have access to this meal-assistance system, I predict that I would use it in my daily life.
  \item \textbf{Perceived Enjoyment:} I find using this meal-assistance system to be enjoyable.
  \item \textbf{Adaptability:} I find this meal-assistance system adaptable to my requirements.
  \item \textbf{Transparency:} I find this meal-assistance system to be transparent.
  \item \textbf{Safety:} I find this meal-assistance system to be safe.
\end{itemize}
\section{Personalization Requests During \\ in-the-wild Evaluation}

\subsection{Meal ID 1: CR1 in Personal Context}

\noindent\textbf{Meal:} Chicken Nuggets and Potato Wedges

\noindent\textbf{Bite Ordering Preference:} Chicken nugget dipped in ranch dipping sauce then potato wedge without dipping

\medskip

\noindent\textbf{Adaptability Request:} Feed me As fast as you can\\
\textbf{Adaptability Response:} The robot's speed for handling drinks and wipes has been successfully increased to make the feeding process faster. However, attempts to speed up the process for using utensils were not successful.

\medskip

\noindent\textbf{Transparency Request:} How can I speed up Feeding me with a utensil\\
\textbf{Transparency Response:} To speed up the process of feeding you with a utensil, you can adjust the ``Speed'' parameter for the relevant behaviors. Specifically, you can change the speed to ``high'' for the following actions:
\begin{itemize}
\item \textbf{PickUtensil}: This will make the robot pick up the utensil faster.
\item \textbf{AcquireBite}: This will speed up the process of picking up the food item.
\item \textbf{TransferUtensil}: This will make the robot bring the utensil to your mouth more quickly.
\item \textbf{StowUtensil}: This will speed up placing the utensil back in the tool mount.
\end{itemize}
You can adjust these settings through the user interface where these parameters are available.

\medskip

\noindent\textbf{Adaptability Request: Increase speed to high when feeding with a utensil}\\
\textbf{Adaptability Response:} The feeding system was updated to increase the speed to high when using a utensil to acquire a bite.

\medskip

\noindent\textbf{Adaptability Request: Skip confirmation screens}\\
\textbf{Adaptability Response:} The robot-assisted feeding system was updated to skip confirmation screens when acquiring a bite with a utensil, transferring a drink, and transferring a wipe. However, skipping the confirmation screen for transferring a utensil was not successful due to an invalid parameter.

\medskip

\noindent\textbf{Adaptability Request: Bring everything as close to my mouth as possible}\\
\textbf{Adaptability Response:} The robot feeding system was adjusted to bring utensils, drinks, and wipes closer to the user's mouth, with a new distance of 0.07 units.

\medskip

\noindent\textbf{Adaptability Request: Use button when completing a transfer when taking a sip}\\
\textbf{Adaptability Response:} The robot-assisted feeding system has been updated to use a button to complete actions when transferring drinks, utensils, and wipes.

\medskip

\noindent\textbf{Transparency Request: What is the default action to complete a transfer}\\
\textbf{Transparency Response:} The default action to complete a transfer is to use a button interaction. This means the robot waits for a button press to confirm that the transfer is complete.

\medskip

\noindent\textbf{Transparency Request: What other ways can I end a transfer besides pushing the button}\\
\textbf{Transparency Response:} Besides pushing the button, you can end a transfer by using the ``sense'' interaction, where the robot detects completion automatically, or by setting it to ``auto\_timeout,'' where the transfer ends after a set time.

\medskip

\noindent\textbf{Adaptability Request: Use the sense interaction when completing a transfer Bite}\\
\textbf{Adaptability Response:} The robot-assisted feeding system has been updated to use sensory feedback when completing the transfer of utensils, drinks, and wipes.

\medskip

\noindent\textbf{Adaptability Request: Use the button to complete a transfer only when taking a sip}\\
\textbf{Adaptability Response:} The system was updated so that a button is now used to complete the transfer when taking a sip, using utensils, or wiping.

\medskip

\noindent\textbf{Adaptability Request: Use sensory to complete bite transactions but not for sips or face wiping}\\
\textbf{Adaptability Response:} The robot feeding system was updated to automatically complete actions for drinking and face wiping, while using sensors to ensure successful bites when using utensils.

\medskip

\noindent\textbf{Adaptability Request: Use sensors to end transaction with utensil use the button to end transactions when taking a sip use the button to end transaction for face wipe}\\
\textbf{Adaptability Response:} The robot-assisted feeding system has been updated to improve how it completes tasks:
\begin{enumerate}
\item When using a utensil, the system now automatically senses when the task is done.
\item When taking a sip or wiping the face, the user needs to press a button to indicate the task is complete.
\end{enumerate}

\medskip

\noindent\textbf{Adaptability Request: Initiate transfers with open mouth gesture}\\
\textbf{Adaptability Response:} The robot-assisted feeding system has been updated to start transferring utensils, drinks, and wipes when it detects an open mouth gesture.

\noindent\textbf{Explanation:} Needed to rephrase and send speed requests to accelerate all tools.\\
\textbf{Pseudo-Intervention:} Had to reset because ``close mouth'' was not enabled (experimenter oversight).\\
\textbf{Intervention (Web Interface):} Web app failed to send information correctly, resulting in no predicted bites.\\
\textbf{Explanation:} Navigating the ``Select Bite'' page to override user commands.\\
\textbf{Intervention (Skill):} Lowered safety limits for outside-mouth transfers from 7cm to 6cm (necessary for CR1).\\
\textbf{Pseudo-Intervention:} Food moved away while CR1 was taking a bite. Stopped the robot but CR1 will attempt manual skill selection.\\
\textbf{Explanation:} Acquisition can fail if the selected keypoint's depth is invalid. Will need to retry.\\
\textbf{Intervention (Hardware):} Drink acquisition failed—robot moved to a seemingly random position. Had to reposition the drink.\\
\textbf{Intervention (Hardware):} Drink acquisition failed again—missed the drink during pickup. Repositioned the drink again.\\
\textbf{Explanation:} Can switch to button control for drink transfer if CR1 prefers.\\
\textbf{Explanation:} Dipping can fail due to unreliable depth detection for liquids. Will need to retry.\\
\textbf{Explanation:} How to use adaptability through prompt engineering for button/sensor-based transfers.\\
\textbf{Intervention (Hardware):} Drink acquisition failed again—robot moved to a random position. Need to check the AR tag and detection.\\
\textbf{Intervention (Web Interface):} Webpage failed to receive plate image. Page didn’t populate and couldn’t recover through manual pickup. Reset and refreshed the page.\\
\textbf{Explanation:} Difference in wording between ``initiate transfer'' and ``end transfer'' in adaptability settings.\\
\textbf{Explanation:} Prompt engineering for alternative bite selection.\\
\textbf{Intervention (Hardware):} Drink acquisition failed again—robot offset to the left and bumped the handle. Reset and retried.

\subsection{Meal ID 2: CR1 Watching TV}

\noindent\textbf{Meal:} Strawberries and whipped cream

\noindent\textbf{Bite Ordering Preference:} Please feed me strawberries dipped and whipped cream

\medskip

\noindent\textbf{Adaptability Request: Dip the strawberry deeper into the whipped cream}\\
\textbf{Adaptability Response:} The system was updated to dip the strawberry deeper into the whipped cream by adjusting the dipping depth to 0.03.

\medskip

\noindent\textbf{Adaptability Request: Be silent}\\
\textbf{Adaptability Response:} The robot-assisted feeding system has been updated to operate silently when transferring drinks, utensils, and wipes.

\medskip

\noindent\textbf{Adaptability Request: Change automatic timer to every five seconds}\\
\textbf{Adaptability Response:} The robot-assisted feeding system was updated to automatically proceed every five seconds when acquiring a bite with a utensil and transferring a utensil or drink. However, the change could not be applied to the wiping action.

\noindent\textbf{Intervention (Hardware):} Strawberry got stuck in whipped cream. Simultaneously, camera anomaly due to CR1 looking away.\\
\textbf{Explanation:} How to get around webpage population error.\\
\textbf{Intervention (Web Interface):} Interface died when trying to finish the meal. Needed to refresh.

\subsection{Meal ID 3: CR2 in Personal Context}

\noindent\textbf{Meal:} Chicken nuggets ketchup and apple slices

\noindent\textbf{Bite Ordering Preference:} Dip chicken nuggets and ketchup and give apple slices after two bites of chicken nuggets

\medskip

\noindent\textbf{Adaptability Request: Don't show continue after picking up apples or chicken nuggets on plate or drink}\\
\textbf{Adaptability Response:} The system was requested to stop showing a ``continue'' prompt after picking up apples, chicken nuggets, or a drink. However, the changes could not be made because the requested settings were not allowed.

\medskip

\noindent\textbf{Adaptability Request: Top show continue confirmation pages on the web interface}\\
\textbf{Adaptability Response:} The robot-assisted feeding system has been updated to ask for confirmation before performing actions like acquiring a bite with a utensil, transferring a drink, and using a wipe.

\medskip

\noindent\textbf{Adaptability Request: Don't show continue pages on the web interface}\\
\textbf{Adaptability Response:} The robot-assisted feeding system has been updated to no longer ask for confirmation when acquiring a bite with a utensil, transferring a drink, or using a wipe.

\medskip

\noindent\textbf{Adaptability Request: Dip chicken and ketchup more}\\
\textbf{Adaptability Response:} The robot's feeding system was adjusted to dip the chicken deeper into the ketchup, enhancing the dipping action.

\medskip

\noindent\textbf{Adaptability Request: Go faster in between picking up bites of apple}\\
\textbf{Adaptability Response:} The robot's speed has been increased when picking up and transferring bites of apple.

\medskip

\noindent\textbf{Adaptability Request: Do not dip apple and ketchup}\\
\textbf{Adaptability Response:} The system was updated to ensure that when using a utensil to serve food, apples are only lightly dipped in sauces like ketchup, with minimal depth.

\noindent\textbf{Explanation:} Can’t go back to bite pickup after dropped bite, what bite ordering should be, and how it can be personalized.\\
\textbf{Explanation:} Difference between continue on web interface and otherwise. CR2 is trying to remove confirmation pages.\\
\textbf{Explanation:} Gesture programming.\\
\textbf{Explanation:} Why it’s dipping apple slices in ketchup, CR2 did not notice that he is selecting ketchup by mistake manually.\\
\textbf{Intervention (Hardware):} Bite stuck in ketchup, had to remove it.\\
\textbf{Explanation:} How to dip more ketchup.\\
\textbf{Intervention (Web Interface):} Webpage not sending message in adaptability, had to refresh web interface.\\
\textbf{Explanation:} Dipping and depth not functioning all the time.

\subsection{Meal ID 4: CR2 in Social Context}

\noindent\textbf{Add Gesture:}\\
Name: Continuous mouth open\\
Description: Mouth open for five seconds\\
Accuracy: 83.33\%

\medskip

\noindent\textbf{Adaptability Request: Show confirmation page before transfer}\\
\textbf{Adaptability Response:} Confirmation pages have been successfully added before transferring drinks and wipes. However, there was an issue adding a confirmation page for utensils.

\medskip

\noindent\textbf{Adaptability Request: Turn off voice}\\
\textbf{Adaptability Response:} The robot's voice has been turned off for all interactions involving drinks, utensils, and wipes.

\medskip

\noindent\textbf{Adaptability Request: Use continuous mouth to detect I'm ready for a drink}\\
\textbf{Adaptability Response:} A user requested a change to the robot's feeding system to recognize when they are ready for a drink using continuous mouth detection. However, there was an issue with the system recognizing the specified command.

\medskip

\noindent\textbf{Adaptability Request: Use continuous mouth open to detect for transfers}\\
\textbf{Adaptability Response:} The robot-assisted feeding system has been updated to use continuous mouth opening as a signal for transferring drinks, utensils, and wipes.

\noindent\textbf{Explanation:} Be silent vs lower voice.\\
\textbf{Explanation:} Use continuous mouth open to show ready for a drink.\\
\textbf{Pseudo-Intervention:} Had to code back continuous transfer in (experimenter fault).\\
\textbf{Intervention (Skill):} Head perception returned none bug.\\
\textbf{Not an Intervention, but:} Went and refilled the mug with water upon CR2 request.\\
\textbf{Intervention (Hardware):} Robot fell while switching controllers after drink transfer, possibly due to a robot minimum voltage error.

\subsection{Meal ID 5: CR2 Watching TV}

\noindent\textbf{Add Gesture:}\\
Name: Head still\\
Description: Hold head still for five seconds\\
Accuracy: 83.33\%

\medskip

\noindent\textbf{Adaptability Request: Head still for transfer completion of mouth wiping but do not make this change for drinking and eating}\\
\textbf{Adaptability Response:} The robot-assisted feeding system was updated to keep the user's head still when wiping their mouth. However, the system will not keep the head still when the user is eating or drinking.

\medskip

\noindent\textbf{Adaptability Request: Be quiet and do not talk at all}\\
\textbf{Adaptability Response:} The robot-assisted feeding system has been updated to operate silently when transferring drinks, utensils, and wipes.

\medskip

\noindent\textbf{Adaptability Request: Bring back confirmation page}\\
\textbf{Adaptability Response:} The system now includes a confirmation step before using utensils to acquire food, transferring drinks, and using wipes.

\noindent\textbf{Intervention:} CR2 Requested that his food be warmed up. Intervention to heat the food in the microwave.\\
\textbf{Explanation:} Auto bites got reset because of ghost clicks.\\
\textbf{Intervention (Web Interface):} Had to refresh webpage to send transfer continue for drink.\\
\textbf{Intervention (Web Interface):} Nose pointer interface CR2 uses stopped working.\\
\textbf{Intervention (Web Interface):} Web interface refresh for pickup bite.

\subsection{Meal ID 6: CR1 in Social Context}

\noindent\textbf{Meal:} Chicken and Hash Browns

\noindent\textbf{Bite Ordering Preference:} Alternate one bite of each

\medskip

\noindent\textbf{Adaptability Request: Move to retract position after every bite}\\
\textbf{Adaptability Response:} A new step was successfully added to the system to move the utensil back after each use. However, this step could not be added for drinks or wipes.

\medskip

\noindent\textbf{Adaptability Request: Increase auto continue time On the webpage to 100 seconds}\\
\textbf{Adaptability Response:} The system was updated to increase the wait time to 100 seconds for acquiring a bite with a utensil and transferring a drink. However, changes could not be made for transferring a wipe due to an issue with the parameter name.

\medskip

\noindent\textbf{Adaptability Request: Be quiet}\\
\textbf{Adaptability Response:} The robot-assisted feeding system has been updated to operate more quietly when transferring utensils, drinks, and wipes.

\noindent\textbf{Explanation:} Auto continue time on the web interface.\\
\textbf{Intervention (Personalization):} Bug in behavior tree param for bite acquisition auto-continue for bite task selection page, manually updated to 100.\\
\textbf{Explanation:} Can move to adaptability page from task selection and back to switch off auto continue.

\subsection{Other Meals}

\noindent\textbf{Training Meal 1:} A training meal conducted with CR1 to explain basic usage of the system. This meal allowed the user to become acquainted with the system.\\

\medskip

\noindent\textbf{Training Meal 2:} A training meal conducted with CR2 to explain basic usage of the system. This meal allowed the user to become acquainted with the system.\\

\medskip

\noindent\textbf{Aborted Meal 1:} A meal was aborted part-way through due to personal time constraints of CR2.\\

\medskip

\noindent\textbf{Aborted Meal 2:} A meal with CR2 was aborted part-way through due to a robot falling error. Our hypothesis for this failure is a voltage drop, reported by the robot firmware, was caused by interference between a router and a power supply. Although the robot remained safe, we determined this error to be significant enough to warrant further testing and mitigation. Thorough testing confirmed the system's safety before we continued. \\

\section{OT Evaluation Details}

\subsection{Evaluation Questions}

\begin{itemize}
    \item I was able to personalize the mealtime-assistance system to my requirements for this scenario (5 point Likert scale ranging from Strongly Disagree to Strongly Agree).
    \item Personalizing the mealtime-assistance system was easy for this scenario (5 point Likert scale ranging from Strongly Disagree to Strongly Agree)
    \item I would choose the personalized version of the system over the default version for this scenario (Yes / No).
\end{itemize}

\subsection{Personalization Requests}

\noindent \emph{Scenario 1:} User has limited head mobility and cannot lean the default amount to take a bite. They prefer outside-mouth transfer over inside-mouth, and thus would still prefer to lean some amount. \\ \textbf{Adaptability Request:} Move robot closer for transfer. \\
\textbf{Adaptability Response:} The robot was moved closer to you when offering utensils, drinks, and wipes.

\noindent \emph{Scenario 2:} User prefers to dip the food item less. \\ \textbf{Adaptability Request:} Dip food item less in ketchup. \\
\textbf{Adaptability Response:} Reduced how far the robot dips food into ketchup.

\noindent \emph{Scenario 3:} User has limited mouth opening and prefers the robot to skewer food with fork tines along the longer axis to avoid wide mouth opening when the fork is horizontal. \\
\textbf{Adaptability Request:} Skewer item vertically. \\
\textbf{Adaptability Response:} The robot feeding system was updated to skewer food vertically.

\noindent \emph{Scenario 4:} User is eating in a personal setting and wants to be done quickly, so they choose to decrease the auto-continue time for bite pickup. \\
\textbf{Adaptability Request:} Decrease auto continue time for bite pick up to five seconds. \\
\textbf{Adaptability Response:} The automatic wait time before picking up a bite has been reduced to five seconds, as requested.

\noindent \emph{Scenario 5:} User is eating in a social setting and prefers using a button for transfer interactions instead of gestures that may interfere with social interaction. \\
\textbf{Adaptability Request:} Used button to initiate and show completion of transfer. \\
\textbf{Adaptability Response:} The robot feeding system now uses a button to start and confirm when handing you utensils, drinks, or wipes.

\noindent \emph{Scenario 6:} User is watching TV while eating and wants to minimize interactions with the web app by removing confirmation pages.\\ 
\textbf{Adaptability Request:} Get rid of confirmation page for bite transfer.\\
\textbf{Adaptability Response:} Removed the confirmation step when transferring a bite, making feeding quicker and smoother.

\noindent \emph{Scenario 7:} User is eating in a personal setting and wants bite transfer to initiate automatically without waiting for user's open-mouth cues. \\
\textbf{Adaptability Request:} Continue initiating bite transfer without my signal. \\
\textbf{Adaptability Response:} The system was updated to automatically continue feeding without waiting for your confirmation signal.\\
\textit{Note: Robot confuses request with removing the confirmation page from the web application.}\\
\textbf{Adaptability Request:} Don't wait for my signal to transfer the bite.\\ \textbf{Adaptability Response:} The robot was updated to automatically feed you each bite without waiting for your confirmation.\\
\textit{Note: Robot confuses again.} \\
\textbf{Transparency Request:} Do I need to signal with opening my mouth to have it transfer a bite or can it automatically transfer with no signal from me. \\ \textbf{Transparency Response:} You need to signal by opening your mouth to let the robot know you're ready for a bite transfer. Currently, the robot does not automatically transfer bites without your signal. \\
\textbf{Adaptability Request:} Auto initiate bite transfer instead of waiting for me to open my mouth. \\
\textbf{Adaptability Response:} The feeding system was updated to automatically start bringing food toward your mouth without waiting for you to open it.

\noindent \emph{Scenario 8:} User wants the mouth wiper to be closer to their mouth, but other tools to remain at their previous distance. \\
\textbf{Adaptability Request:} Bring wipe 3 cm from my face before transfer.\\
\textbf{Adaptability Response:} You requested bringing the wipe closer (3 cm) to your face before transferring, but this distance was outside the allowed safety limits. The system did not accept this change. \\
\textbf{Transparency Request:} How close can you bring the wipe to my mouth before transfer.\\
\textbf{Transparency Response:} I can bring the wipe as close as 15 centimeters (about 6 inches) from your mouth before you start cleaning. \\
\textbf{Adaptability Request:} Bring the wipe as close as possible to my mouth before transfer. \\
\textbf{Adaptability Response:} The wipe will now be brought closer to your mouth before transferring it.

\noindent \emph{Scenario 9:} User prefers the robot to move quickly. \\
\textbf{Adaptability Request:} Make the robot as fast as possible. \\
\textbf{Adaptability Response:} The robot's movements for picking up, using, and putting away utensils, drinks, and wipes have all been set to the fastest possible speed.

\noindent \emph{Scenario 10:} User is eating in a social setting and the robot's bite acquisition and transfer configurations occlude their social partner. They want the robot to reposition to a non-occluding configuration between bites. \\
\textbf{Adaptability Request:} Retract after every bite transfer. \\
\textbf{Adaptability Response:} The robot now moves back after every bite it gives you.

\end{document}